\documentclass{article}

\usepackage[numbers, sort&compress]{natbib}

\usepackage[main, final]{neurips_2026}

\usepackage[utf8]{inputenc} 
\usepackage[T1]{fontenc}    
\usepackage{hyperref}       
\usepackage{url}            
\usepackage{booktabs}       
\usepackage{amsfonts}       
\usepackage{nicefrac}       
\usepackage{microtype}      
\usepackage{xcolor}         

\usepackage{bm}
\usepackage{wrapfig}
\usepackage{enumitem}
\usepackage{placeins}

\newcommand{\blue}[1]{{\color{blue}#1}}

\usepackage{array}
\usepackage{algorithm}
\usepackage{algpseudocode}
\usepackage{amsmath}
\usepackage{booktabs}
\usepackage{tabularx}
\usepackage{adjustbox}
\usepackage{multirow}
\usepackage{graphicx}
\usepackage{float}
\usepackage{colortbl}
\usepackage{makecell}
\usepackage{caption}
\usepackage{relsize}
\usepackage{marvosym}
\usepackage{arydshln}
\usepackage{hhline}

\definecolor{lightyellow}{RGB}{255,255,200}
\definecolor{lightorange}{RGB}{255,230,200}
\definecolor{lightred}{RGB}{255,200,200}
\definecolor{projectblue}{RGB}{0,102,204}

\newcommand{\tbest}[1]{\cellcolor{lightyellow}#1}
\newcommand{\sbest}[1]{\cellcolor{lightorange}#1}
\newcommand{\best}[1]{\cellcolor{lightred}#1}

\newcolumntype{P}[1]{>{\centering\arraybackslash}m{#1}}

\renewcommand\addcontentsline[3]{}

\title{Any 3D Scene is Worth 1K Tokens: 3D-Grounded Representation for Scene Generation at Scale}

\author{%
  \textbf{Dongxu Wei}$^{1,2,*}$ \quad
  \textbf{Qi Xu}$^{1,*}$ \quad
  \textbf{Zhiqi Li}$^{1,3}$ \quad
  \textbf{Hangning Zhou}$^{2,\dagger}$ \quad
  \textbf{Cong Qiu}$^{2}$ \\
  \textbf{Hailong Qin}$^{2}$ \quad
  \textbf{Mu Yang}$^{2}$ \quad
  \textbf{Zhaopeng Cui}$^{3}$ \quad
  \textbf{Peidong Liu}$^{1}$\textsuperscript{\Letter} \\
  \normalfont
  $^{1}$Westlake University \quad
  $^{2}$Afari Intelligent Drive \quad
  $^{3}$Zhejiang University \\
  $^{*}$Equal contribution. \quad
  $^{\dagger}$Project Lead. \quad
  \textsuperscript{\Letter}Corresponding Author. \\
  Project Page: \href{https://wswdx.github.io/3DRAE}{\textcolor{projectblue}{\texttt{wswdx.github.io/3DRAE}}}
}

\begin{document}

\maketitle

\begin{figure}[!htbp]
\centering
\vspace{-20pt}
\includegraphics[width=0.9\linewidth]{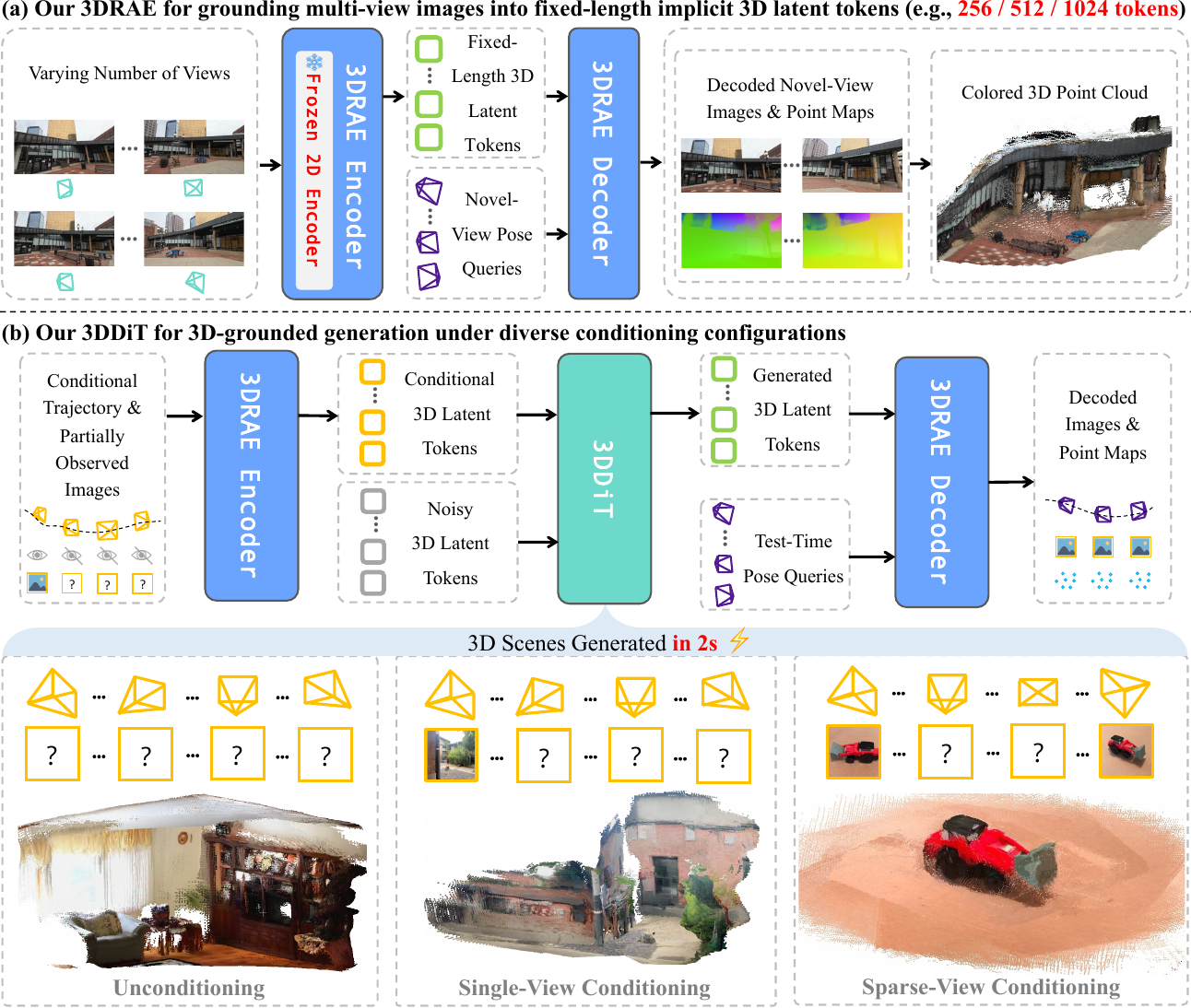}
\vspace{0pt}
\caption{\textbf{3D-Grounded Scene Generation.} (a) Our 3DRAE repurposes frozen 2D foundation models to ground varying number of views into fixed-length implicit 3D latent tokens, which can be queried to decode images and point maps of various views including novel views. (b) Our 3DDiT performs diffusion modeling directly in this implicit 3D latent space, enabling 3D-grounded generation with superior efficiency and spatial consistency.}
\label{fig:teaser}
\vspace{0pt}
\end{figure}

\begin{abstract}
Scene-level 3D generation has long been dominated by 2D multi-view or video diffusion models. Typically, these 2D-based approaches perform scene generation in a 2D latent space, which introduces two fundamental issues: (i) representing 3D scenes via 2D views leads to significant representation redundancy, and (ii) latent space rooted in 2D inherently limits the spatial consistency of the generated scenes.
In this paper, we propose to perform 3D scene generation directly within a high-dimensional implicit 3D latent space derived from powerful 2D foundation models (e.g., DINOv2, SigLIP2, DA3).
Then we tame diffusion transformers to perform diffusion modeling directly in the 3D latent space, enabling 3D-grounded scene generation.
In our experiments, we not only demonstrate the advantages of our method over 2D-based approaches in terms of efficiency and spatial consistency, but also comprehensively analyze the impact of 3D latent representation on 3D scene generation performance. Furthermore, we validate the favorable scalability of our method with respect to representation and model capacity.
\end{abstract}

\section{Introduction}
If we paraphrase the recent literature by characterizing ``the evolution of generative modeling as a continual redefinition of where and how models learn to represent data''~\cite{rae}, then in the domain of 3D generation, this exploration specifically focuses on how models learn to represent 3D data. Early methodologies attempted to leverage 3D priors such as triplanes~\cite{eg3d,gaudi,ssdnerf}, voxels~\cite{platonicgan,hologan,blockgan,holodiffusion} or neural radiance field~\cite{graf,giraffe,nfldm} for 3D representation. Recently, with the advancement of video diffusion models~\cite{ldm,svd,videodiff,cogvideox,hunyuanvideo,videocrafter2,wan}, it is observed that these models can synthesize high-quality video sequences.
This led to a paradigm shift: treating spatial extrapolation as a special case of temporal extension. By adapting video diffusion models—which inherently possess a prior for 2D temporal coherence—to the task of spatially-consistent multi-view image synthesis, these approaches achieve coherent generation of multi-view images that follow a long camera trajectory. This capability enables the simulation of navigation through 3D environments, and has thus become the prevailing approach in most contemporary 3D scene generation frameworks~\cite{cat3d,director3d,prometheus,motionctrl,splatflow,wonderworld,wonderland,voyager,seva,viewcrafter,flexworld,vmem}.

\begin{wrapfigure}{lt}{0.5\textwidth}
    \vspace{-15pt}
  \begin{center}
    \includegraphics[width=\linewidth]{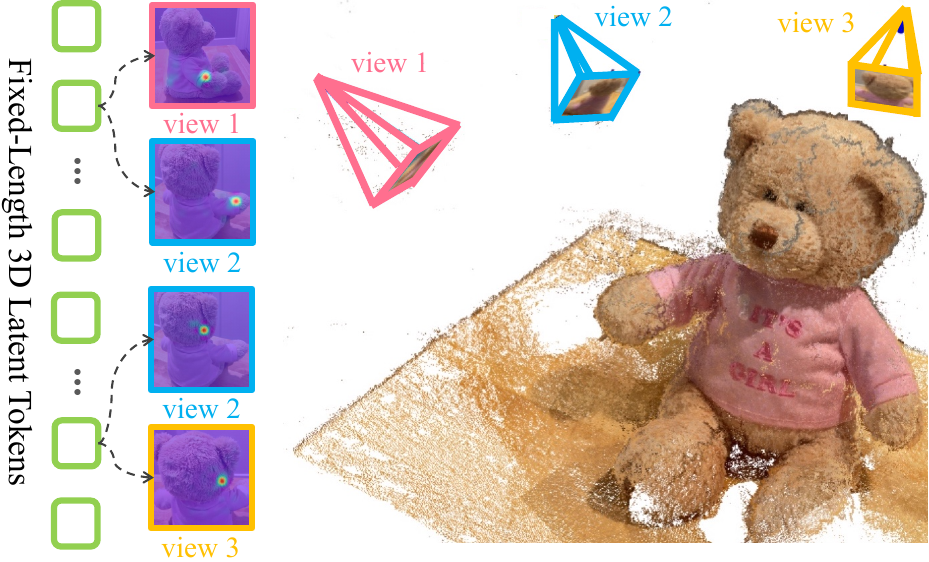}
  \end{center}
\vspace{-5pt}
\caption{We empirically find that overlapping regions across multiple views correspond to the same set of tokens. Therefore, we employ fixed-length 3D latent tokens to eliminate such multi-view redundancy.}
    \label{fig:tokens}
    \vspace{-13pt}
\end{wrapfigure}
In essence, however, these methods remain rooted in 2D latent space rather than constituting truly 3D-grounded generation. The degradation of a 3D scene into 2D views introduces two fundamental issues: (1) \emph{Representation Redundancy}: Significant cross-view overlap inherent in multi-view representations leads to substantial information redundancy.
For instance, with a patch size of 16, depicting a 3D scene using 32 images at 256$\times$256 resolution requires 32$\times$16$\times$16 = 8192 tokens, whereas the actual number of unique informative tokens may be as low as 1024 due to cross-view overlap (Fig.\ref{fig:tokens}), incurring considerable computational waste.
(2) \emph{Limited Spatial Consistency}: As 2D diffusion models lack intrinsic 3D spatial modeling, the 3D scenes they generate through temporal extension inevitably suffer from spatial inconsistencies, where errors may accumulate as pose changes and finally lead to structural deterioration.
Consequently, we pose a critical question: Given only data access to multi-view images or videos (as in previous 2D-based approaches), \emph{is it possible to develop models that learn to transform 2D images into a 3D latent space? And can we perform diffusion modeling directly within the 3D latent space for 3D scene generation?} We conjecture that such 3D‑grounded generation may not need to sacrifice efficiency or spatial consistency.

Motivated by this, our foremost task is to define a 3D latent space that can be derived from 2D images and grounded with 3D knowledge. One approach is to borrow from feed‑forward 3D reconstruction methods, where novel view rendering~\cite{ibrnet,lgm,pixelsplat,gslrm,mvsplat,depthsplat,noposplat,omniscene,siu3r} or point reconstruction~\cite{dust3r,mast3r,vggt,da3} serves as supervision, enabling the model to recover 3D information from 2D images. However, most of these methods rely on pixel‑aligned representations coupled with 2D views, which inevitably leads to representation redundancy. Very few approaches~\cite{lvsm,cut3r,rayzer} attempt to map 2D images into a fixed‑size set of implicit tokens, achieving decoupling from 2D views while remaining grounded in 3D.
However, directly leveraging these high-dimensional tokens (e.g., 768-dim) to construct a latent space amenable to diffusion modeling is challenging and non‑trivial, as diffusion models are generally known to struggle with denoising in high‑dimensional spaces.
Inspired by recent advances in representation autoencoders (RAE)~\cite{rae,rae2} that achieve single image diffusion directly in high‑dimensional spaces with rich semantics, we propose 3DRAE — a method that repurposes pre‑trained 2D foundation models (e.g., DINOv2~\cite{dinov2}, SigLIP2~\cite{siglip2}, DA3~\cite{da3}) to encode varying number of 2D views into fixed‑length implicit and 3D-grounded latent tokens (Fig.\ref{fig:teaser}-a).
Then we introduce 3DDiT — a 3D-grounded diffusion transformer that performs 3D scene generation directly in the 3D latent space learned by our 3DRAE (Fig.\ref{fig:teaser}-b), enabling 3D-grounded generation with significantly improved efficiency and spatial consistency compared to prior 2D diffusion-based methods.
Furthermore, we thoroughly investigated how 3D latent representations derived from different 2D foundation models impact 3D scene generation, and achieved a trade-off between reconstruction and generation capabilities across various design choices.

In summary, our main contributions are as follows:
\begin{itemize}[leftmargin=1em,itemsep=0pt,topsep=0pt,parsep=2pt]
    \item To our knowledge, this is the first work to extend single‑image RAE to the 3D domain and thoroughly discuss the impact of 3D latent representation on 3D scene generation.
    \item Extensive experiments demonstrate the significant potential of our method over 2D diffusion‑based methods, particularly in terms of efficiency and spatial consistency.
    \item Our method exhibits strong scalability in terms of both representation and model capacity, and we provide models of varying complexity to benefit the community.
\end{itemize}
\section{Related Work}
\noindent\textbf{Scene-Level 3D Generation.}
With the advancement of open-source video diffusion models~\cite{svd,cogvideox,hunyuanvideo,wan}, many 3D generation methods~\cite{cat3d,director3d,prometheus,motionctrl,splatflow,wonderworld,wonderland,voyager,seva,viewcrafter,flexworld,vmem,gen3r} have emerged that incorporate camera conditions to synthesize videos following specified trajectories. These approaches effectively degrade 3D spatial extrapolation to 2D temporal extension. To improve spatial consistency, some methods~\cite{director3d,prometheus,splatflow} additionally predict 3DGS primitives alongside RGB during decoding, enabling 3D-aware synthesis. Others~\cite{viewcrafter,flexworld,vmem} maintain explicit 3D memory based on historical frames, and condition subsequent generation on historical renderings to enforce coherence.
Despite these efforts, all such methods fundamentally operate in a 2D latent space, which introduces substantial representation redundancy and struggles to ensure spatial consistency at the 3D level.
Another line of work~\cite{hunyuan3d,trellis} focuses on object-level 3D generation. Leveraging massive high-quality 3D asset data, these methods achieve 3D generation in explicit 3D latent space based on point clouds or voxels. However, for scene-level generation, such high-quality 3D data is unavailable. How to achieve 3D-grounded scene generation using only multi-view image data remains a challenge.

\noindent\textbf{Representation for 2D Diffusion.}
Recent work in 2D diffusion reveals that aligning the latent space with pre-trained representation encoders substantially improves 2D generation performance. Some approaches~\cite{vavae,maetok,dcae,ldetok} align VAE latents with pretrained encoders or incorporate MAE~\cite{mae} objectives into VAE training, significantly enhancing reconstruction quality. Others align intermediate diffusion features with pretrained encoders~\cite{repa,ddt,reg} or generate the features directly within a diffusion model~\cite{redi,rae,rae2}, accelerating convergence. Among these, Representation Autoencoder (RAE)~\cite{rae,rae2} pioneers the use of frozen 2D foundation models as autoencoders for diffusion and generates high-dimensional latent tokens with rich semantics directly. This not only improves training stability and generation fidelity but also enables visual understanding and generation to operate within a shared representation, opening new possibilities for unified vision models.
Inspired by RAE, we propose 3DRAE, which lifts 2D representations into a 3D latent space for 3D-grounded generation with superior efficiency and spatial consistency.
\section{Method}
\vspace{-0.5em}
\subsection{Preliminaries}
\textbf{Autoencoders for Diffusion.}
Diffusion models operate through an encode-decode paradigm that maps data to a latent space and subsequently learn to generate samples within this latent space. Prior image or video diffusion approaches~\cite{ldm,svd,hunyuanvideo,cogvideox,wan} typically employ channel-wise compressed VAE encoder $\mathcal{E}$ to map images into the 2D latent space, and use decoder $\mathcal{D}$ to reconstruct images from latents. For an image $I \in \mathbb{R}^{H \times W \times 3}$, the VAE pipeline follows:
\begin{figure}[!ht]
\centering
\vspace{-5pt}
\includegraphics[width=0.67\linewidth]{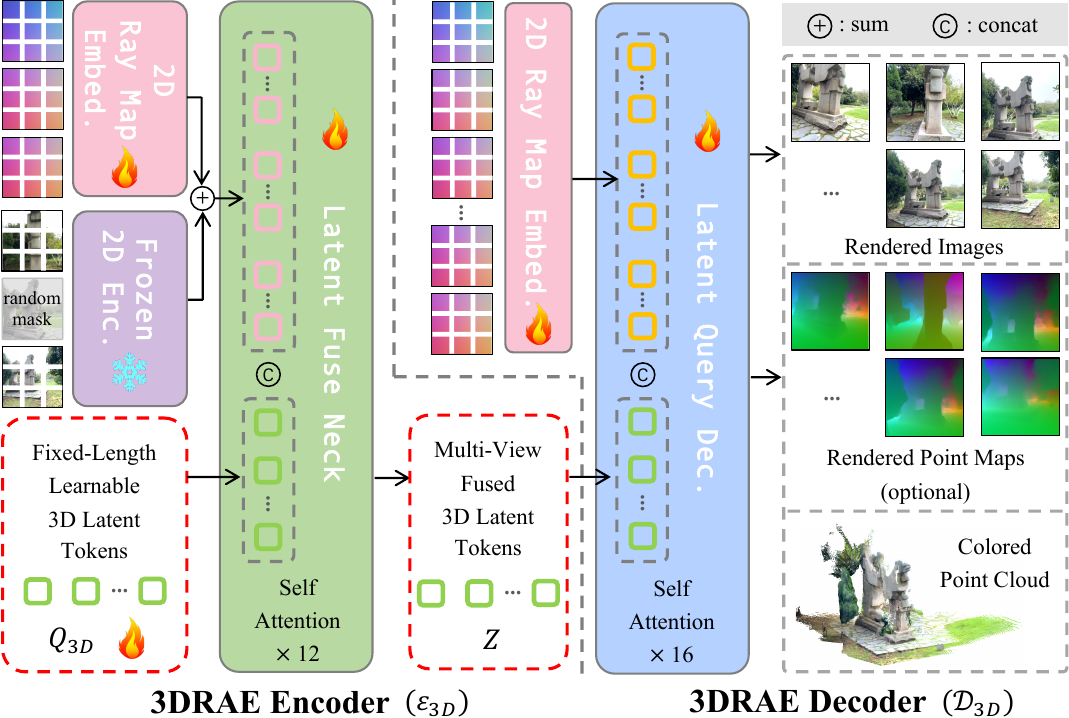}
\vspace{0pt}
\caption{\textbf{The Architecture of Our 3DRAE.}}
\label{fig:3drae}
\vspace{-15pt}
\end{figure}
\begin{small}
\begin{equation}
    I \xrightarrow{\mathcal{E}} z \xrightarrow{\mathcal{D}} \hat{I}, \quad \text{where } z = \mathcal{E}(I) \in \mathbb{R}^{h \times w \times c},
\end{equation}
\end{small}with spatial compression factor $f$ such that $h=H/f, w=W/f$, and channel dimension $c$ (e.g., $c=4$). While efficient, such VAEs yield low-capacity latent optimized primarily for reconstruction, lacking the global semantic structure essential for robust generalization and generative performance.

{\noindent}A recent line of work~\cite{rae,rae2} introduces representation autoencoders (RAEs) to reformulate this pipeline using frozen visual encoders $f_{\text{enc}}$ (e.g., DINOv2~\cite{dinov2}, SigLIP2~\cite{siglip2}), which are primarily pretrained for semantic understanding. For an image $I$, the RAE mapping becomes:
\begin{small}
\begin{equation}
    I \xrightarrow{f_{\text{enc}}} z \xrightarrow{\mathcal{D}} \hat{I}, \quad \text{where } z = f_{\text{enc}}(I) \in \mathbb{R}^{n \times d},
\end{equation}
\end{small}with patch size $p$ such that the number of patch tokens $n = \frac{H}{p} \times \frac{W}{p}$, and feature dimension $d$ (e.g., $d=768$). Unlike VAEs, RAEs retain rich semantics without channel compression, significantly accelerating convergence of diffusion training and improving image fidelity. However, since the latent space of RAEs remains grounded at 2D image level, they cannot be used to represent 3D scenes.

\subsection{3DRAE: 3D-Grounded Latent Representation Autoencoder}
In this work, we demonstrate how to leverage pre-trained 2D foundation models to extract multi-view features, which are subsequently transformed into a 3D-grounded latent space endowed with rich 2D semantic information.
Our approach encodes multi-view observations of a 3D scene into a compact representation consisting of fixed-length 3D latent tokens, from which various viewpoints including novel ones can be decoded, effectively decoupling 3D scene representation from input view observations.

\noindent\textbf{3D Scene Encoding and Decoding.}
Given a set of $N_{\text{obs}}$ observed multi-view images $\mathcal{I}_{\text{obs}} = \{I_i\}_{i=1}^{N_{\text{obs}}}$ with corresponding camera poses $\mathcal{P}_{\text{obs}} = \{P_i\}_{i=1}^{N_{\text{obs}}}$, our 3DRAE encoder $\mathcal{E}_\text{3D}$ establishes the mapping:
\begin{small}
\begin{equation}
    (\mathcal{I}_\text{obs}, \mathcal{P}_{\text{obs}}) \xrightarrow{\mathcal{E}_{\text{3D}}} Z, \quad \text{where } Z = \mathcal{E}_{\text{3D}}(\mathcal{I}_\text{obs}, \mathcal{P}_{\text{obs}}) \in \mathbb{R}^{m \times d}.
\end{equation}
\end{small}Here $Z$ denotes 3D latent tokens of fixed length $m$, which compactly encodes the scene appearance and geometry. Notably, $N_{\text{obs}}$ can vary to accommodate representation of 3D scenes observed by varying number of views.

{\noindent}As for decoding, our 3DRAE decoder $\mathcal{D}_\text{3D}$ is designed to recover appearance and geometry for varying views given the encoded 3D latent tokens $Z$. For example, for a camera trajectory $\mathcal{P_\text{dec}} = \{P_j\}_{j=1}^{N_{\text{dec}}}$ consisting of $N_{\text{dec}}$ views to be decoded, the view-conditioned decoding then follows:
\begin{small}
\begin{equation}
    (Z, \mathcal{P_\text{dec}}) \xrightarrow{\mathcal{D}_\text{3D}} \{\hat{I}_j\}_{j=1}^{N_{\text{dec}}}, \quad \text{where } \hat{I}_j = \mathcal{D}_\text{3D}(Z, \psi(P_j)) \in \mathbb{R}^{H \times W \times 3}.
\end{equation}
\end{small}Here $\psi(P_j)$ generates ray map tokens that query $Z$ to retrieve appearance and geometry for the specified view pose $P_j$. This formulation treats the 3D latent $Z$ as a sufficient scene statistic: once encoded from partial observations, it enables rendering of the full scene given any camera pose along the trajectory $\mathcal{P}_{\text{obs}}$, effectively transforming partial observations into a complete 3D representation.

\noindent\textbf{Model Design.}
As illustrated in Fig.\ref{fig:3drae}, our 3DRAE encoder $\mathcal{E}_\text{3D}$ incorporates a frozen 2D image encoder $f_{\text{enc}}$ and a 2D ray map embedding module $\psi$ to extract multi-view visual and geometric information, respectively. Their outputs are summed to form a dense set of multi-view patch tokens:
\begin{small}
\begin{equation}
    t_i = f_{\text{enc}}(I_i) + \psi(P_i) \in \mathbb{R}^{n \times d}, \quad \text{for } i = 1, \dots, N_{\text{obs}}.
\end{equation}
\end{small}Subsequently, we introduce a Latent Fuse Neck, which concatenates a fixed-length set of learnable 3D latent tokens $Q_Z \in \mathbb{R}^{m \times d}$ with the multi-view patch tokens $\{t_i\}_{i=1}^{N_{\text{obs}}}$. The combined tokens are processed through a 12-layer transformer with self attention, enabling deep cross-modal information fusion. The output is a set of multi-view fused 3D latent tokens $Z \in \mathbb{R}^{m \times d}$ that encapsulate comprehensive 3D scene information after spatial redundancy reduction.

{\noindent}Next, we employ a latent query decoder $\mathcal{D}_\text{3D}$ consisting of a 16-layer transformer with self attention to integrate ray map tokens $\psi(P_j)$ of target views with the encoded 3D latent tokens $Z$, where the ray map tokens query and retrieve relevant information for synthesizing the target views. The resulting ray map tokens from the final decoder layer are then unpatchified to image renderings. Meanwhile, another latent query decoder with DPT head~\cite{dpt} is optionally employed to produce 3D point maps for the target views, where the DPT head processes multi-layer ray map tokens to produce high-resolution point maps.

\noindent\textbf{Training.}
To ensure our 3D latent space is decoupled from multi-view inputs and maintains view invariance, we employ a training strategy that samples data with varying resolutions, aspect ratios, and view counts $N_{\text{obs}}$. This enables our 3D latent tokens to represent 3D scenes under flexible observational configurations.

{\noindent}As for supervision, in addition to supervising rendered images with MSE and perceptual losses, we introduce an adversarial loss following common practices in 2D VAEs~\cite{vaegan}. For the optional point map head, we adopt an aleatoric-uncertainty loss to weight the discrepancy between the predicted point map and ground truth. The overall training objective is:
\begin{small}
\begin{equation}
    \mathcal{L} = \text{MSE}(\hat{\mathbf{I}}^t, \mathbf{I}^t) + \lambda_1 \cdot \text{LPIPS}(\hat{\mathbf{I}}^t, \mathbf{I}^t) + \lambda_2 \cdot \omega_G \cdot \text{GAN}(\hat{\mathbf{I}}^t, \mathbf{I}^t) + \lambda_3 \cdot \mathcal{L}_{pmap}(\hat{\mathbf{P}}^t,\hat{\mathbf{C}}^t,\mathbf{P}^t),
    \label{Eq:loss_stage1}
\end{equation}
\end{small}where $\hat{\mathbf{I}}^t$ and $\mathbf{I}^t$ are the predicted and ground-truth images. $\hat{\mathbf{P}}^t$ and $\mathbf{P}^t$ denote the predicted and ground-truth point maps. $\hat{\mathbf{C}}^t$ denotes the uncertainty map. The definition of $\mathcal{L}_{pmap}$ follows VGGT~\cite{vggt}.


\subsection{3DDiT: Taming Diffusion Transformers for 3D-Grounded Generation}
\label{3ddit}
With the proposed 3DRAE capable of mapping any multi-view observed 3D scene into a compact 3D latent space, we now turn to modeling this space with a 3D diffusion transformer (3DDiT) for 3D-grounded generation. Our 3DDiT effectively learns to transform empty or partial 2D observations into complete 3D scene realizations by denoising in the 3D latent space.

\noindent\textbf{Model Design.}
The architecture of our 3DDiT is shown in Fig.\ref{fig:teaser}-b. Following~\cite{rae}, we adopt the LightningDiT~\cite{lightningdit} backbone with wide diffusion head, scaling the DiT~\cite{dit} width to match the high dimensionality of our 3D latent tokens.
Unlike prior methods that sample from 2D latent tokens proportional to the number of views, our 3DDiT operates on the compact, fixed-length 3D latent tokens. Since the 3D latent space are comprehensively grounded in 3D via novel view synthesis and 3D reconstruction in our 3DRAE, it inherently yields better efficiency and spatial consistency.

\noindent\textbf{Diverse Conditioning Configurations.}
Since our 3DRAE encoder can map varying number of images and their poses to the 3D latent space, we can leverage it to encode conditioning signals from observational views under diverse configurations for our 3DDiT.
Particularly, we decompose conditioning into: (i) spatial extent specified by camera trajectory $\mathcal{P}_{\text{cond}} = \{P_i\}_{i=1}^{N_{\text{pose}}}$, and (ii) appearance specified by multi-view images $\mathcal{I}_{\text{cond}} = \{I_j\}_{j=1}^{N_{\text{img}}}$, where $N_\text{pose}\geq N_\text{img}$ since the camera trajectory should comprehensively covers the scene's spatial extent, while multi-view image observations may be partial or absent. As shown in Fig.\ref{fig:teaser}-b, we define three conditioning configurations:
\begin{itemize}[leftmargin=1em,itemsep=0pt,topsep=0pt,parsep=2pt]
    \item \textbf{Unconditioning} ($N_{\text{img}}=0$): Only the camera trajectory $\mathcal{P}_\text{cond}$ is provided, defining the spatial scope without appearance constraints. The scene should be generated from scratch.
    \item \textbf{Single-view conditioning} ($N_{\text{img}}=1$): In addition to the trajectory $\mathcal{P}_\text{cond}$, one reference image from a particular viewpoint is provided to guide appearance, demanding strong spatial completion to unobserved regions.
    \item \textbf{Sparse-view conditioning} ($N_{\text{img}}>1$): In addition to the trajectory $\mathcal{P}_\text{cond}$, several images from multiple viewpoints are provided for appearance reference, requiring faithful reconstruction and cross-view spatial consistency.
\end{itemize}

{\noindent}To accommodate the case where $N_\text{pose} > N_\text{img}$, we introduce a random masking strategy (see details in \ref{randmask}) when training 3DRAE to make it aware that some of the input views are invisible.

\noindent\textbf{Flow Matching in 3D Latent Space.}
We adopt the flow matching framework~\cite{flowmatching,rectifiedflow} to model the probability path of our 3D latent space. Given a clean latent $Z_0 \in \mathbb{R}^{m \times d}$ encoded by 3DRAE and a condition $c$ comprising the partial observations $\mathcal{I}_\text{cond}$ and camera trajectory $\mathcal{P}_\text{cond}$, we construct a linearly interpolated probability path between $Z_0$ and Gaussian noise $\epsilon \sim \mathcal{N}(0, \mathbf{I})$:
\begin{small}
\begin{equation}
    Z_t = (1-t) Z_0 + t \epsilon, \quad \text{where } t \sim \mathcal{U}[0,1],\quad Z_t\in\mathbb{R}^{m\times d}.
\end{equation}
\end{small}The associated velocity field is $u_t = \epsilon - Z_0$. Our 3DDiT $v_\theta$ learns to predict this velocity conditioned on the noisy latent $Z_t$, timestep $t$, and condition $c$:
\begin{small}
\begin{equation}
    \mathcal{L}_{\text{FM}} = \mathbb{E}_{t, Z_0, \epsilon} \left[ \left\| v_\theta(Z_t, t, c) - (\epsilon - Z_0) \right\|_2^2 \right].
\end{equation}
\end{small}
\section{Experiments}
In this section, we present necessary implementation details in § \ref{implementation_detail}, benchmark settings in § \ref{bench}, as well as baseline and metric settings in § \ref{baseline_metric}.
Then we evaluate our method across various settings to demonstrate its effectiveness in § \ref{main_res}. We also present ablations and discussions in § \ref{discuss}. \textbf{Additional experimental details and results can be found in our Appendix and Supplementary Videos.}

\subsection{Implementation Details}
\label{implementation_detail}
Our 3DRAE is trained on 32 A800 GPUs for 160k steps, using an 8k-step warm-up and a peak learning rate of 2$\times10^{-4}$. Our 3DDiT is trained for 80k steps with the same hardware and learning rate schedule.
To process multi-view image data, we follow large 3D reconstruction models\cite{da3,pi3} and adopt a dynamic data sampling strategy, where the batch size is dynamically determined by the sampled resolution, aspect ratio, and number of views.
In particular, for each GPU, the image resolution is randomized to fit the total pixel count between 37,000 and 75,500, the aspect ratio is randomly sampled from the range [0.7, 1.8], and the total number of views ranges from 6 to 18.
During the training of 3DRAE, we randomly mask out input views with a probability of 0.1 and supervise the model solely with novel-view image and point map rendering.
During the training of 3DDiT, we randomly mask out input views with a probability ranging from 0.6 to 0.9 to simulate varying conditional view counts, and mask out all views with a probability of 0.1 to accommodate unconditional generation. Such training strategy enables our method to support diverse conditioning configurations.
\begin{table*}[!ht]
    \vspace{-10pt}
    \centering
    \footnotesize
    \setlength{\tabcolsep}{3pt} 

    \begin{minipage}{0.9\linewidth}
        \centering
        \begin{adjustbox}{max width=\linewidth}
        \begin{tabular}{c l c *{5}{c} *{5}{c}}
        \toprule
        \multicolumn{1}{c}{} & \multirow{2}{*}{Method} & \multirow{2}{*}{Runtime$\downarrow$} &
        \multicolumn{5}{c}{DL3DV~\cite{dl3dv}} &
        \multicolumn{5}{c}{RE10K~\cite{re10k}} \\
        \cmidrule(lr){4-8} \cmidrule(lr){9-13}
        & & & PSNR$\uparrow$ & LPIPS$\downarrow$ & FID$\downarrow$ & $\text{ATE}_\text{r}\downarrow$ & $\text{ATE}_\text{t}\downarrow$ & PSNR$\uparrow$ & LPIPS$\downarrow$ & FID$\downarrow$ & $\text{ATE}_\text{r}\downarrow$ & $\text{ATE}_\text{t}\downarrow$ \\
        \midrule
        \multirow{7}{*}{\rotatebox[origin=c]{90}{\textbf{1-view}}} &
        DepthSplat\cite{depthsplat} & {\best{0.15s}} & {9.11} & {0.631} & {141.14} & {1.27} & {0.155} & {11.65} & {0.584} & {113.03} & {0.85} & {0.102} \\
        & LVSM\cite{lvsm} & {\sbest{0.21s}} & \tbest{14.58} & {0.581} & {114.83} & {1.64} & {0.197} & \best{17.42} & \best{0.385} & \tbest{30.94} & \tbest{0.55} & {0.077} \\
        & MotionCtrl\cite{motionctrl} & {11.8s} & {10.72} & {0.655} & {128.77} & {2.32} & {0.290} & {10.23} & {0.672} & {115.55} & {2.47} & {0.260} \\
        & ViewCrafter\cite{viewcrafter} & {200.4s} & {9.31} & {0.667} & {94.08} & {1.58} & {0.196} & {13.77} & {0.454} & {55.88} & {1.08} & {0.110} \\
        & SEVA\cite{seva} & {85.2s} & \sbest{14.60} & \sbest{0.511} & \sbest{45.96} & \sbest{0.32} & \sbest{0.029} & \tbest{15.52} & \tbest{0.426} & \sbest{27.21} & \sbest{0.28} & \best{0.021} \\ 
        & 3DRAE (ours) & {\tbest{0.23s}} & {14.07} & \tbest{0.520} & \tbest{84.67} & \tbest{1.33} & \tbest{0.127} & {14.95} & {0.466} & {30.48} & {0.80} & \tbest{0.070} \\
        & 3DDiT (ours) & \underline{1.98s} & \best{14.62} & \best{0.480} & \best{40.93} & \best{0.29} & \best{0.025} & \sbest{15.77} & \sbest{0.411} & \best{24.67} & \best{0.26} & \best{0.021} \\
        \midrule
        \multirow{7}{*}{\rotatebox[origin=c]{90}{\textbf{2-view}}} &
        DepthSplat\cite{depthsplat} & \best{0.16s} & {13.57} & {0.637} & {146.06} & {1.41} & {0.171} & {16.17} & {0.369} & {51.98} & {0.43} & {0.042} \\
        & LVSM\cite{lvsm} & \sbest{0.23s} & \best{17.52} & {0.468} & {89.85} & {1.32} & {0.151} & \best{23.22} & \best{0.180} & \sbest{17.59} & \tbest{0.26} & {0.022} \\
        & MotionCtrl\cite{motionctrl} & {\texttt{N/A}} & {\texttt{N/A}} & {\texttt{N/A}} & {\texttt{N/A}} & {\texttt{N/A}} & {\texttt{N/A}} & {\texttt{N/A}} & {\texttt{N/A}} & {\texttt{N/A}} & {\texttt{N/A}} & {\texttt{N/A}} \\
        & ViewCrafter\cite{viewcrafter} & {202.2s} & {12.57} & {0.560} & {117.94} & {1.59} & {0.200} & {15.36} & {0.429} & {57.31} & {0.83} & {0.089} \\
        & SEVA\cite{seva} & {86.4s} & \tbest{16.22} & \tbest{0.433} & \sbest{35.96} & \sbest{0.31} & \sbest{0.033} & \tbest{19.44} & \tbest{0.260} & {18.30} & {0.28} & \tbest{0.021} \\
        & 3DRAE (ours) & \tbest{0.24s} & \sbest{17.19} & \best{0.374} & \tbest{37.07} & \tbest{0.47} & \tbest{0.060} & \sbest{19.57} & \sbest{0.256} & \best{15.99} & \sbest{0.19} & \sbest{0.019} \\
        & 3DDiT (ours) & \underline{2.01s} & {15.95} & \sbest{0.415} & \best{33.15} & \best{0.27} & \best{0.023} & {18.19} & {0.298} & \tbest{17.70} & \best{0.18} & \best{0.015} \\
        \bottomrule
        \end{tabular}
        \end{adjustbox}
    \end{minipage}

    \vspace{0.1cm}

    \begin{minipage}{0.9\linewidth}
        \centering
        \begin{adjustbox}{max width=\linewidth}
        \begin{tabular}{c l c *{5}{c} *{5}{c}}
        \toprule
        \multicolumn{1}{c}{} & \multirow{2}{*}{Method} & \multirow{2}{*}{Runtime$\downarrow$} &
        \multicolumn{5}{c}{CO3D~\cite{co3d}} &
        \multicolumn{5}{c}{ScanNet~\cite{scannet}} \\
        \cmidrule(lr){4-8} \cmidrule(lr){9-13}
        & & & PSNR$\uparrow$ & LPIPS$\downarrow$ & FID$\downarrow$ & $\text{ATE}_\text{r}\downarrow$ & $\text{ATE}_\text{t}\downarrow$ & PSNR$\uparrow$ & LPIPS$\downarrow$ & FID$\downarrow$ & $\text{ATE}_\text{r}\downarrow$ & $\text{ATE}_\text{t}\downarrow$ \\
        \midrule
        \multirow{7}{*}{\rotatebox[origin=c]{90}{\textbf{1-view}}} &
        DepthSplat\cite{depthsplat} & \best{0.15s} & {8.07} & {0.671} & {153.80} & {1.72} & \tbest{0.220} & {10.19} & {0.607} & {141.85} & \tbest{0.43} & {0.119} \\
        & LVSM\cite{lvsm} & \sbest{0.21s} & {15.32} & {0.616} & {192.31} & {2.22} & {0.303} & \sbest{15.34} & \tbest{0.500} & {87.19} & {0.72} & {0.172} \\
        & MotionCtrl\cite{motionctrl} & {11.8s} & {11.35} & {0.673} & {124.90} & {2.25} & {0.432} & {10.04} & {0.672} & {149.80} & {2.09} & {0.325} \\
        & ViewCrafter\cite{viewcrafter} & {200.4s} & {9.31} & {0.667} & {152.32} & {2.23} & {0.427} & {12.69} & {0.533} & {79.91} & {1.40} & {0.232} \\
        & SEVA\cite{seva} & {85.2s} & \best{17.31} & \sbest{0.462} & \sbest{42.50} & \sbest{0.92} & \sbest{0.029} & {14.47} & {0.507} & \sbest{53.11} & \sbest{0.28} & \sbest{0.065} \\ 
        & 3DRAE (ours) & \tbest{0.23s} & \tbest{15.43} & \tbest{0.536} & \tbest{113.4} & \tbest{1.60} & {0.248} & \tbest{15.24} & \sbest{0.472} & \tbest{66.93} & {0.59} & \tbest{0.117} \\
        & 3DDiT (ours) & \underline{1.98s} & \sbest{17.01} & \best{0.460} & \best{42.29} & \best{0.68} & \best{0.026} & \best{15.74} & \best{0.448} & \best{47.97} & \best{0.22} & \best{0.062} \\
        \midrule
        \multirow{7}{*}{\rotatebox[origin=c]{90}{\textbf{2-view}}} &
        DepthSplat\cite{depthsplat} & \best{0.16s} & {15.74} & {0.616} & {146.01} & {1.71} & {0.281} & {15.31} & {0.472} & {104.97} & {0.70} & {0.114} \\
        & LVSM\cite{lvsm} & \sbest{0.23s} & {16.87} & {0.569} & {167.60} & {2.19} & {0.359} & \best{20.04} & \sbest{0.323} & {54.89} & {0.33} & {0.087} \\
        & MotionCtrl\cite{motionctrl} & {\texttt{N/A}} & {\texttt{N/A}} & {\texttt{N/A}} & {\texttt{N/A}} & {\texttt{N/A}} & {\texttt{N/A}} & {\texttt{N/A}} & {\texttt{N/A}} & {\texttt{N/A}} & {\texttt{N/A}} & {\texttt{N/A}} \\
        & ViewCrafter\cite{viewcrafter} & {202.2s} & {10.24} & {0.664} & {174.71} & {2.16} & {0.414} & {14.03} & {0.492} & {79.76} & {1.36} & {0.218} \\
        & SEVA\cite{seva} & {86.4s} & \best{19.96} & \best{0.375} & \tbest{36.95} & \tbest{0.92} & \tbest{0.028} & \tbest{18.48} & \tbest{0.348} & \sbest{36.75} & \tbest{0.26} & \tbest{0.069} \\ 
        & 3DRAE (ours) & \tbest{0.24s} & \sbest{19.34} & \sbest{0.382} & \best{34.18} & \sbest{0.73} & \best{0.022} & \sbest{19.61} & \best{0.314} & \best{36.73} & \best{0.21} & \sbest{0.050} \\
        & 3DDiT (ours) & \underline{2.01s} & \tbest{18.21} & \tbest{0.411} & \sbest{34.33} & \best{0.68} & \sbest{0.025} & {18.28} & {0.355} & \tbest{38.73} & \best{0.21} & \best{0.048} \\
        \bottomrule
        \end{tabular}
        \end{adjustbox}
    \end{minipage}
    \vspace{-0.0pt}
    \caption{\textbf{Quantitative Comparison.} We compare both generation-oriented (1-view) and reconstruction-oriented (2-view) settings with state-of-the-art regression-based (DepthSplat, LVSM) and diffusion-based (MotionCtrl, ViewCrafter, SEVA) methods. 2-view setting is not available ($\texttt{N/A}$) for MotionCtrl that only supports 1-view conditioning. PSNR and LPIPS are used for \textbf{\emph{reconstruction-prioritized evaluation}}. FID is used for \textbf{\emph{generation-prioritized evaluation}}. $\text{ATE}_\text{r}$ and $\text{ATE}_\text{t}$ are used for \textbf{\emph{spatial consistency evaluation}}.}
    \label{tab:main_res}
    \vspace{-2em}
\end{table*}
Our 3DRAE and 3DDiT are trained on a large collection of 17 multi-view datasets (Table \ref{tab:supl:datasets}), spanning $\sim$130K scenes ranging from indoor to outdoor and from object-centric to general environments. Please refer to our Appendix for further implementation and data processing details.

\subsection{Benchmark}
\label{bench}
To comprehensively evaluate our method across diverse settings, we collected 358 scenes in total from four datasets—DL3DV\cite{dl3dv}, RealEstate10K\cite{re10k}, CO3D\cite{co3d} and ScanNet\cite{scannet}. This collection covers a wide range of scenarios, from indoor to outdoor and from object-centric to general scenes.
For each scene, we sample a trajectory from the original video that covers the main subject with substantial camera motion. We evaluate methods by predicting unseen views given only one or few conditional views. To assess performance under varying conditioning settings, we randomly sample different numbers of conditional views from the trajectory as reference. Fewer conditional views correspond to sparser observations, shifting the evaluation toward 3D generation capability; more conditional views correspond to denser observations emphasize 3D reconstruction fidelity.

\begin{figure}[!ht]
\centering
\vspace{-2em}
\includegraphics[width=1.0\linewidth]{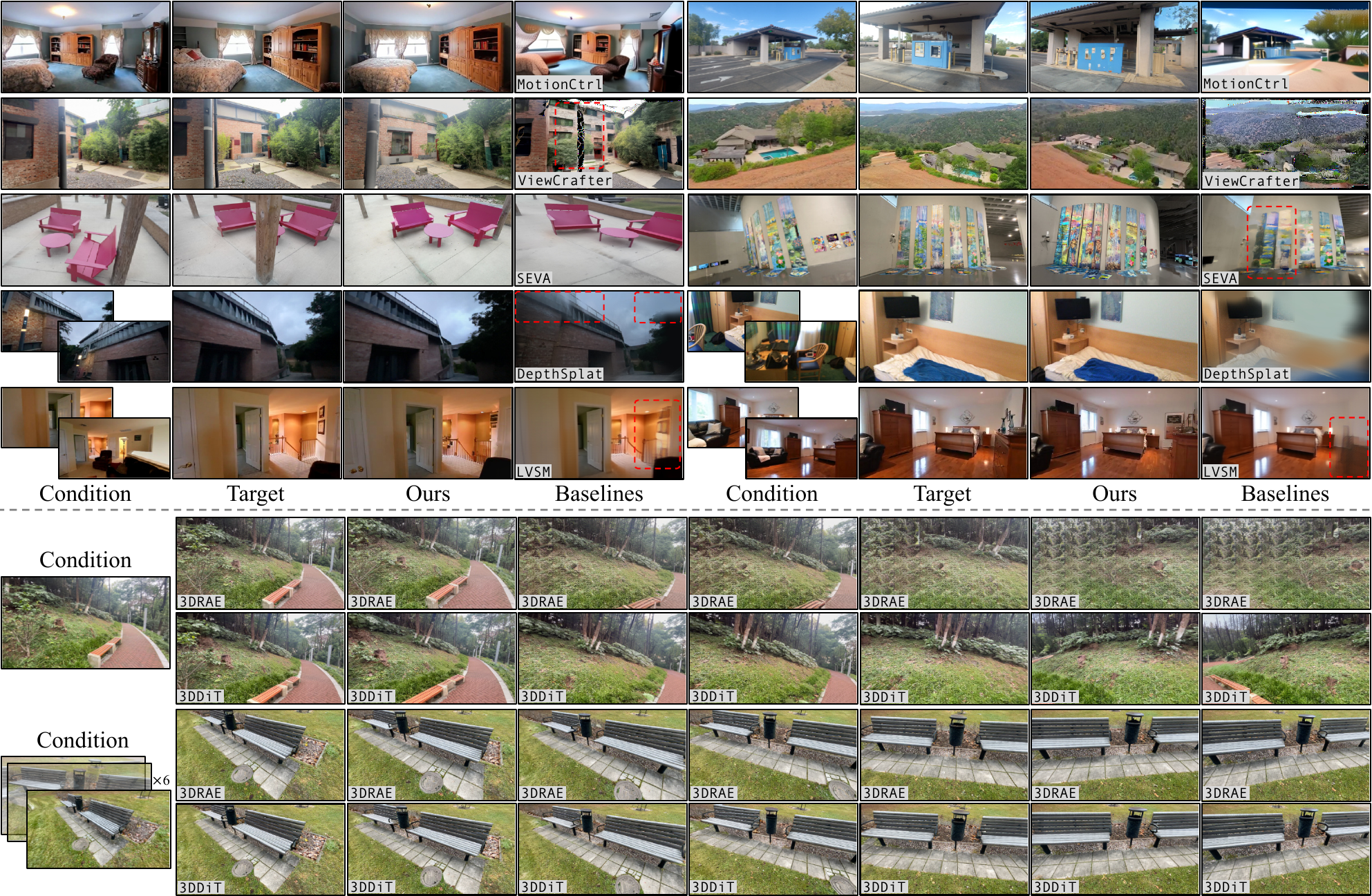}
\vspace{-8pt}
\caption{\textbf{Qualitative Comparison.} \textbf{Top}: Comparison with state-of-the-art methods. \textbf{Bottom}: Comparison between our 3DRAE and 3DDiT to demonstrate the effectiveness of diffusion modeling, where 3DRAE fails in 1-view setting that prioritizes generation capability. The red bounding boxes highlight failure cases in baselines.}
\label{fig:main_comp}
\vspace{-2em}
\end{figure}
\subsection{Baselines and Metrics}
\label{baseline_metric}
\noindent\textbf{Baselines.}
We compare our method against both regression-based (DepthSplat\cite{depthsplat}, LVSM\cite{lvsm}) and diffusion-based approaches (MotionCtrl\cite{motionctrl}, ViewCrafter\cite{viewcrafter}, SEVA\cite{seva}). Regression-based methods excel at faithfully reconstructing 3D scenes under dense observation. Diffusion-based methods, by contrast, leverage the generative capabilities of video diffusion models and are better suited for completing unseen regions under partial observation. All methods follow their official implementations and are evaluated on our benchmark.

\noindent\textbf{Metrics.}
\textbf{\emph{For reconstruction quality}}, we adopt PSNR and LPIPS to measure photometric and perceptual differences between predicted and ground-truth images. \textbf{\emph{For generation quality}}, we use FID to quantify the distance between generated and real image distributions. \textbf{\emph{To assess spatial consistency}}, we follow prior work~\cite{viewcrafter} to employ pose accuracy, where $\text{ATE}_\text{r}$ and $\text{ATE}_\text{t}$ denote rotation and translation errors between predicted and ground-truth camera poses.
We note that PSNR and LPIPS are indicative only when scene observations are relatively complete (i.e., number of conditional views > 1). When conditioned on a single view, generated images increasingly involve content synthesized from scratch as the camera moves. In such cases, FID for image quality and pose accuracy for spatial consistency become more appropriate evaluation focuses.

\subsection{Main Results}
\label{main_res}
\noindent\textbf{Quantitative Comparison.}
Our quantitative results are presented in Table \ref{tab:main_res}.
All methods are evaluated under both single-view and sparse-view conditioning settings. Runtime is computed as the average time cost to render 16 views, which exhibits minimal variation across different conditioning settings for all methods.
Notably, our 3DDiT achieves substantially shorter runtime compared to all 2D diffusion-based approaches. This efficiency stems from denoising directly in a compact 3D latent space, which eliminates the representation redundancy introduced by 2D views.
Our 3DRAE also matches regression-based methods in runtime.
In terms of performance, when only single view is available for conditioning (1-view), our 3DDiT demonstrates clear superiority compared to all baselines across all datasets and metrics, especially in FID and pose accuracy, highlighting its advantages in generation quality and spatial consistency.
For sparse-view conditioning (2-view, start-end-view) where reconstruction fidelity rather than generative diversity is prioritized, our 3DRAE sometimes outperforms 3DDiT. This is because 3DDiT is not designed for faithful reconstruction. The denoising process of diffusion transformer may cause generated scenes to deviate from conditional observations.
Overall, the results show that both our 3DRAE and 3DDiT achieve competitive performance under sparse-view conditioning, demonstrating the effectiveness of our method.

\noindent\textbf{Qualitative Comparison.}
We also present qualitative results in Fig.\ref{fig:main_comp} for more intuitive demonstration.
\emph{For 2D diffusion-based methods} (i.e., MotionCtrl\cite{motionctrl}, ViewCrafter\cite{viewcrafter}, SEVA\cite{seva}), we focus on comparing generation quality and thus show results under single-view conditioning.
Under large motion, MotionCtrl misaligns with target poses. ViewCrafter uses point cloud renderings as condition signals but suffers black holes from occluded regions. SEVA adds dense anchor frames, improving quality and adherence, yet images still deviate, blur, or have artifacts. In contrast, our 3DDiT consistently produces high-quality images that align accurately with target views.
\emph{For regression-based methods} (i.e., DepthSplat\cite{depthsplat}, LVSM\cite{lvsm}), the focus is reconstruction quality. We therefore present results with two views as conditioning. Despite overlap in conditional views, both DepthSplat and LVSM fail to handle large motion and occlusions, as they lack generative capabilities.
In contrast, our method plausibly completes the occluded regions with superior rendering quality.

{\noindent}In the bottom of Fig.\ref{fig:main_comp}, we further compare 3DRAE and 3DDiT under both generation-prioritized (1-view) and reconstruction-prioritized (6-view) settings. In 1-view setting, 3DRAE fails to extrapolate to unobserved regions, with quality deteriorating as viewpoint changes. 3DDiT, benefiting from diffusion modeling, generates plausible occluded regions while preserving spatial consistency. In 6-view setting, since the conditioning appearance is complete, both models perform well.

\subsection{Discussions}
\label{discuss}
\noindent\textbf{Choices for 2D Foundation model.}
Since our 3DRAE derives 3D latent tokens from 2D representation models, we explore the impact on 3D reconstruction and generation by varying the 2D foundation model of 3DRAE and further yielding variants of 3DDiT.
We also implement the variant (3DRAE-ImgOnly, 3DDiT-ImgOnly) without any representation model for reference, where we replace multi-view features derived from representation models with raw image patch embeddings in 3DRAE.
\begin{table*}[!ht]
    \vspace{-5pt}
    \centering
    \footnotesize
    \setlength{\tabcolsep}{3pt} 

    \begin{minipage}{0.4\linewidth}
        \centering
        \begin{adjustbox}{max width=\linewidth}
        \begin{tabular}{l *{2}{c} *{2}{c}}
        \toprule
        \multirow{2}{*}{Split / 6-view} &
        \multicolumn{2}{c}{DL3DV~\cite{dl3dv}} &
        \multicolumn{2}{c}{RE10K~\cite{re10k}} \\
        \cmidrule(lr){2-3} \cmidrule(lr){4-5}
        & PSNR$\uparrow$ & LPIPS$\downarrow$ & PSNR$\uparrow$ & LPIPS$\downarrow$ \\
        \midrule
        3DRAE-ImgOnly & \sbest{21.42} & \sbest{0.226} & \sbest{25.49} & \sbest{0.124} \\
        3DRAE-DA3\cite{da3} & \best{21.62} & \best{0.221} & \best{26.38} & \best{0.118} \\
        3DRAE-SigLIP2\cite{siglip2} & \tbest{19.40} & {0.283} & \tbest{22.60} & {0.187} \\
        3DRAE-DINOv2\cite{dinov2} & {19.22} & \tbest{0.281} & {21.98} & \tbest{0.186} \\
        \bottomrule
        \end{tabular}
        \end{adjustbox}
    \end{minipage}
    \begin{minipage}{0.477\linewidth}
        \centering
        \begin{adjustbox}{max width=\linewidth}
        \begin{tabular}{l *{3}{c} *{3}{c}}
        \toprule
        \multirow{2}{*}{Split / 1-view} &
        \multicolumn{3}{c}{DL3DV~\cite{dl3dv}} &
        \multicolumn{3}{c}{RE10K~\cite{re10k}} \\
        \cmidrule(lr){2-4} \cmidrule(lr){5-7}
        & FID$\downarrow$ & $\text{ATE}_\text{r}\downarrow$ & $\text{ATE}_\text{t}\downarrow$ & FID$\downarrow$ & $\text{ATE}_\text{r}\downarrow$ & $\text{ATE}_\text{t}\downarrow$ \\
        \midrule
        3DDiT-ImgOnly & {51.68} & {1.02} & {0.093} & {30.31} & {0.89} & {0.047} \\
        3DDiT-DA3\cite{da3} & \tbest{47.75} & \tbest{0.54} & \tbest{0.057} & \tbest{29.76} & \tbest{0.66} & \tbest{0.034} \\
        3DDiT-SigLIP2\cite{siglip2} & \sbest{44.65} & \sbest{0.32} & \sbest{0.029} & \sbest{26.74} & \sbest{0.29} & \sbest{0.024} \\
        3DDiT-DINOv2\cite{dinov2} & \best{40.93} & \best{0.29} & \best{0.025} & \best{24.67} & \best{0.26} & \best{0.021} \\
        \bottomrule
        \end{tabular}
        \end{adjustbox}
    \end{minipage}

    \vspace{0.0pt}
    \caption{\textbf{Ablation on 2D Foundation Model.}}
    \label{tab:abl_2denc}
    \vspace{-2pt}
\end{table*}
\begin{table*}[!ht]
    \vspace{-2pt}
    \centering
    \footnotesize
    \setlength{\tabcolsep}{3pt} 

    \begin{minipage}{0.45\linewidth}
        \centering
        \begin{adjustbox}{max width=\linewidth}
        \begin{tabular}{l c *{2}{c} *{2}{c}}
        \toprule
        \multirow{2}{*}{Split / 6-view} & \multirow{2}{*}{Runtime$\downarrow$} &
        \multicolumn{2}{c}{DL3DV~\cite{dl3dv}} &
        \multicolumn{2}{c}{RE10K~\cite{re10k}} \\
        \cmidrule(lr){3-4} \cmidrule(lr){5-6}
        & & PSNR$\uparrow$ & LPIPS$\downarrow$ & PSNR$\uparrow$ & LPIPS$\downarrow$ \\
        \midrule
        3DRAE-tok256 & \best{0.10s} & \tbest{18.22} & \tbest{0.324} & \tbest{20.73} & \tbest{0.226} \\
        3DRAE-tok512 & \sbest{0.17s} & \sbest{18.76} & \sbest{0.301} & \sbest{21.36} & \sbest{0.207} \\
        3DRAE-tok1024 & \tbest{0.23s} & \best{19.22} & \best{0.281} & \best{21.98} & \best{0.186} \\
        \bottomrule
        \end{tabular}
        \end{adjustbox}
    \end{minipage}
    \begin{minipage}{0.52\linewidth}
        \centering
        \begin{adjustbox}{max width=\linewidth}
        \begin{tabular}{l c *{3}{c} *{3}{c}}
        \toprule
        \multirow{2}{*}{Split / 1-view} & \multirow{2}{*}{Runtime$\downarrow$} &
        \multicolumn{3}{c}{DL3DV~\cite{dl3dv}} &
        \multicolumn{3}{c}{RE10K~\cite{re10k}} \\
        \cmidrule(lr){3-5} \cmidrule(lr){6-8}
        & & FID$\downarrow$ & $\text{ATE}_\text{r}\downarrow$ & $\text{ATE}_\text{t}\downarrow$ & FID$\downarrow$ & $\text{ATE}_\text{r}\downarrow$ & $\text{ATE}_\text{t}\downarrow$ \\
        \midrule
        3DDiT-XL-tok256 & \tbest{1.41s} & \tbest{42.02} & \tbest{0.31} & \best{0.025} & \tbest{25.47} & \sbest{0.28} & \sbest{0.022} \\
        3DDiT-XL-tok512 & {1.67s} & \sbest{41.85} & \sbest{0.30} & \sbest{0.027} & \sbest{25.12} & \sbest{0.28} & \best{0.021} \\
        3DDiT-S-tok1024 & \best{1.14s} & {44.83} & {0.33} & {0.031} & {28.31} & {0.46} & {0.027} \\
        3DDiT-B-tok1024 & \sbest{1.35s} & {42.83} & {0.32} & {0.029} & {26.83} & {0.38} & {0.025} \\
        3DDiT-XL-tok1024 & {1.98s} & \best{40.93} & \best{0.29} & \best{0.025} & \best{24.67} & \best{0.26} & \best{0.021} \\
        \bottomrule
        \end{tabular}
        \end{adjustbox}
    \end{minipage}
    \vspace{0.1cm}

    \vspace{-5pt}
    \caption{\textbf{Ablation on Representation and Model Capacity.}}
    \label{tab:abl_cap}
    \vspace{-10pt}
\end{table*}
For 3DRAE that prioritizes reconstruction, we evaluate the impact of 2D foundation model under 6-view conditioning, where observations are sufficiently dense. For 3DDiT that focuses on generation, we conduct analysis under 1-view conditioning, where most scene content need to be generated from scratch.

\begin{figure}[!ht]
    \centering
        \vspace{-5pt}
        \includegraphics[width=0.9\textwidth]{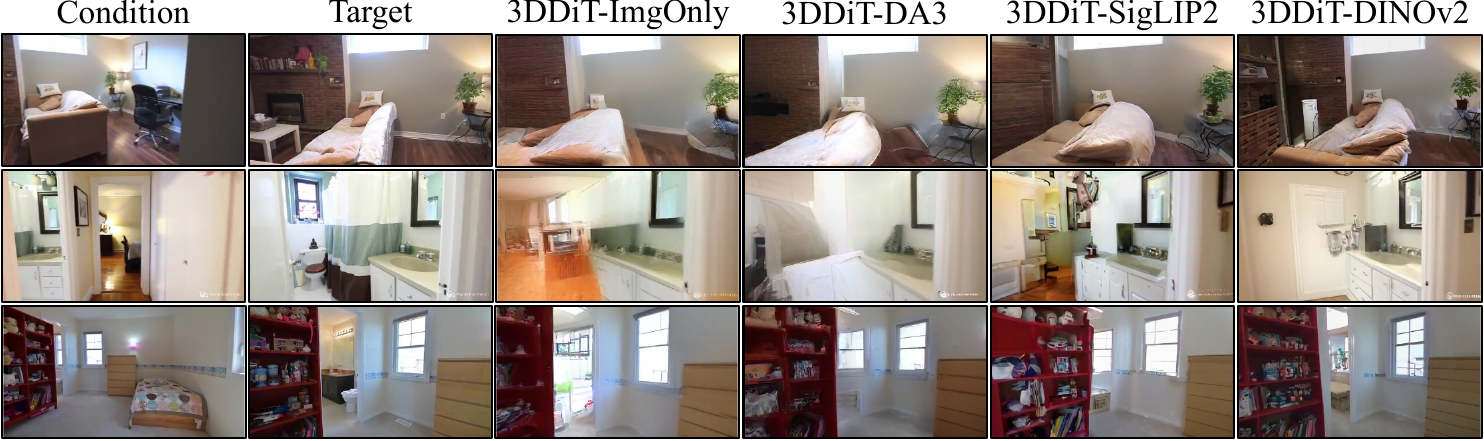}
        \vspace{-5pt}
        \caption{\textbf{Qualitative Ablations on 2D Foundation Models.}}
        \label{fig:enc_comp}
        \vspace{0pt}
\end{figure}

{\noindent}We report the results on DL3DV and RE10K datasets in Table \ref{tab:abl_2denc}.
Among all 3DRAE variants, the one built upon DA3\cite{da3} achieves the best reconstruction performance, demonstrating that 3D prior knowledge from large multi-view reconstruction models benefits the reconstruction capability of 3D latent representation. In contrast, 3DRAEs built upon semantics-prioritized 2D representations (i.e., DINOv2\cite{dinov2}, SigLIP2\cite{siglip2}) exhibits worse reconstruction fidelity, even underperforming the image-only baseline without any representation model. This implies that \emph{high-level semantic representation is deficient in faithful reconstruction}, consistent with findings in 2DRAEs\cite{rae} where variants based on DINOv2/SigLIP2 underperform those based on reconstruction-focused models like MAE\cite{mae}.
For 3DDiT, the results reveals a striking reversal.
Surprisingly, variants built upon DINOv2 and SigLIP2 achieve substantially better generation quality and spatial consistency than their DA3-based and image-only counterparts, suggesting that \emph{semantic priors are crucial for enhancing 3D generation}. Qualitative results in Fig.\ref{fig:enc_comp} further support this finding: DA3-based and image-only 3DDiTs fail to learn general scene appearance, wheareas DINOv2 and SigLIP2-based models generate plausible appearance.
Since this work is centered on 3D generation, we default to DINOv2 as our 2D encoder for the best generation performance.

\noindent\textbf{Scaling with Capacity.}
We further evaluate the scalability of both 3DRAE and 3DDiT with respect to representation capacity (i.e., the number of 3D latent tokens) and model capacity (i.e., the size of 3DDiT model). Following the same protocol as Table \ref{tab:abl_2denc}, we report reconstruction and generation results under their respective evaluation settings in Table \ref{tab:abl_cap}.
We can see that while reducing the number of 3D latent tokens significantly degrades 3DRAE reconstruction fidelity, the resulting more compact 3D latent space benefits diffusion modeling—such that the generation quality does not deteriorate sharply with fewer tokens.
Notably, \emph{our 3DDiT with 256 tokens achieves highly competitive generation performance while offering substantially better efficiency.}
Furthermore, generation quality scales positively with 3DDiT model size, demonstrating the scalability of our method with the number of parameters.

\begin{table}[!hb]
    \vspace{0pt}
    \centering
    \setlength{\tabcolsep}{10pt}
    \resizebox{0.45\linewidth}{!}{
    \begin{tabular}{lcc}
    \toprule
    \multirow{2}{*}{Split / 1-view} & DL3DV~\cite{dl3dv} & RE10K~\cite{re10k} \\
    \cmidrule(lr){2-2} \cmidrule(lr){3-3}
    & FID$\downarrow$ & FID$\downarrow$ \\
    \midrule
    3DDiT-w/o IN & 43.15 & 26.15 \\
    3DDiT-w/ IN & \textbf{40.93} & \textbf{24.67} \\
    \bottomrule
    \end{tabular}
    }
    \vspace{5pt}
    \caption{\textbf{Ablation on ImageNet (IN) pre-training.}}
    \label{tab:abl_in}
    \vspace{-15pt}
\end{table}
\noindent\textbf{Compatibility with Single-Image Data.}
\label{exp_imgnet}
Since our 3DRAE can lift an arbitrary number of views---including a single image---to the 3D latent space, we can leverage single-image data during 3DDiT pre-training as described in § \ref{3ddit}.
As shown in Table \ref{tab:abl_in}, benefiting from the diversity and scale of ImageNet~\cite{imagenet}, pre-training equips our 3DDiT with general scene appearance priors and improves generation quality.
This highlights the compatibility of our approach with single-image data and its potential to scale to massive 2D image and video datasets.

\FloatBarrier
\section{Conclusion}
In this paper, we propose a novel approach that performs diffusion modeling directly within an compact and implicit 3D latent space. Our method grounds semantic representations from frozen 2D foundation models with explicit 3D awareness, transforming 2D features that scale linearly with the number of views into view-invariant 3D latent tokens.
A diffusion transformer is then built upon this 3D latent space, enabling 3D scene generation with superior efficiency and spatial consistency compared to prior 2D-based approaches.
Furthermore, we comprehensively investigate the impact of 2D foundation model on 3D generation performance and reveal that semantic prior knowledge substantially benefits generation quality. Finally, we validate the scalability of our model with respect to both representation capacity and model size, demonstrating the strong potential of our approach for scaling up to broader model regimes.

\noindent\textbf{Limitations.}
\label{limitation}
In this paper, we train our model on as many large-scale multi-view datasets as we can access, and demonstrate the strong generalization ability of the proposed 3D latent representation for out-of-distribution 3D scene reconstruction and generation. Our method also shows significant potential to outperform 2D diffusion-based approaches in both efficiency and spatial consistency. Nevertheless, compared to the 2D image and video generation models developed by well-resourced industrial groups---which are trained on web-scale image and video data, often supplemented with large amounts of proprietary in-house data---the scale of our training data remains relatively limited. To fully unlock the potential of 3D latent representation for 3D scene generation, it is necessary to further scale up the data volume by leveraging massive Internet image and video data, thereby improving the fundamental capabilities of the model.

\begin{ack}
This work was supported in part by NSFC under Grant 62202389, in part by a grant from the Westlake University-Muyuan Joint Research Institute, in part by the Westlake Education Foundation, and in part by the Open Project Program of the State Key Laboratory of CAD\&CG (Grant No. A2513), Zhejiang University.
\end{ack}

{
    \small
    \bibliographystyle{unsrt}
    \bibliography{main}
}

\clearpage

\appendix

\section*{Appendix}
\label{appendix}
\section{3DRAE Implementation}
The overall architecture detail and hyperparameter settings for all 3DRAE variants are presented in Table \ref{tab:supl:3drae_detail}.
In general, our 3DRAE is trained on a large collection of multi-view datasets for 120k steps, the discriminator is not trained until 50k steps, and the adversarial loss is not enabled until 60k steps.
By default, we use DINOv2\cite{dinov2} as the frozen 2D encoder in our 3DRAE for all experiments if not otherwise specified.
The hyperparameters for the dynamic data sampling and random masking strategy are presented in Table \ref{tab:supl:3drae_detail}(b).
In the following sections, we describe the settings in detail for each module separately.
\begin{table}[ht!]
\vspace{-5pt}
\centering
\resizebox{0.8\linewidth}{!}{
\begin{tabular}{lll}
\toprule
\multicolumn{3}{c}{\textbf{(a) Network Architecture}} \\ \hline
\specialrule{0em}{1pt}{0pt}
\multirow{13}{*}{\textbf{Frozen 2D Encoder \qquad\qquad}} & \multicolumn{2}{l}{\textbf{DepthAnything3 (DA3)~\cite{da3}}} \qquad\qquad\qquad\qquad\qquad\qquad\\
\multirow{13}{*}{} & Model Config\qquad\qquad\qquad\qquad\qquad\qquad & ViT-Base \\ 
\multirow{13}{*}{} & Patch Size & 14 \\
\multirow{13}{*}{} & Hidden Dim & 768 \\ 
\cmidrule(r){2-3}
\multirow{13}{*}{} & \multicolumn{2}{l}{\textbf{DINOv2~\cite{dinov2}}} \\
\multirow{13}{*}{} & Model Config & ViT-Base \\ 
\multirow{13}{*}{} & Patch Size & 14 \\
\multirow{13}{*}{} & Hidden Dim & 768 \\ 
\cmidrule(r){2-3}
\multirow{13}{*}{} & \multicolumn{2}{l}{\textbf{SigLIP2~\cite{siglip2}}} \\
\multirow{13}{*}{} & Model Config & ViT-Base \\ 
\multirow{13}{*}{} & Patch Size & 16 \\
\multirow{13}{*}{} & Hidden Dim & 768 \\ 
\hline
\specialrule{0em}{1pt}{0pt}
\multirow{5}{*}{\textbf{Ray Map Embedding}} & Input Channel & 6+1=7\\
\multirow{5}{*}{} & Patch Size & 14 \\
\multirow{5}{*}{} & Embed Dim & 768 \\
\multirow{5}{*}{} & Conv Kernel & 14 \\ 
\multirow{5}{*}{} & Conv Stride & 14 \\
\hline
\specialrule{0em}{1pt}{0pt}
\multirow{5}{*}{\textbf{Latent Fuse Neck}} & Depth & 12\\
\multirow{5}{*}{} & Num Latent Tokens & 256/512/1024 \\ 
\multirow{5}{*}{} & Hidden Dim & 768 \\
\multirow{5}{*}{} & Num Heads & 16 \\ 
\multirow{5}{*}{} & MLP Ratio & 4 \\ 
\hline
\specialrule{0em}{1pt}{0pt}
\multirow{6}{*}{\textbf{Latent Query Decoder}} & Depth & 16\\
\multirow{6}{*}{} & Patch Size & 14 \\ 
\multirow{6}{*}{} & Hidden Dim & 768 \\
\multirow{6}{*}{} & Num Heads & 16 \\ 
\multirow{6}{*}{} & MLP Ratio & 4 \\ 
\multirow{6}{*}{} & DPT Layer Index & [3, 7, 11, 15] \\
\hline
\toprule
\specialrule{0em}{3pt}{0pt}
\multicolumn{3}{c}{\textbf{(b) Hyperparameters}} \\
\specialrule{0em}{1pt}{0pt} \hline
\specialrule{0em}{1pt}{1pt}
\textbf{Loss Weights} & $\lambda_1,\lambda_2,\lambda_3$ & 1.0, 0.75, 1.0 \\ \hline
\specialrule{0em}{1pt}{0pt}
\multirow{15}{*}{\textbf{Training Details}} & lr schedule & Cosine Decay \\
\multirow{15}{*}{} & lr & 2e-4 \\
\multirow{15}{*}{} & Warmup Steps & 8000 \\
\multirow{15}{*}{} & Optimizer & AdamW \\
\multirow{15}{*}{} & $\beta_1$, $\beta_2$ & 0.9, 0.95 \\
\multirow{15}{*}{} & Gradient Clip & 1.0 \\
\multirow{15}{*}{} & Iterations & 120000 \\
\multirow{15}{*}{} & Disc Start Iter & 50000 \\
\multirow{15}{*}{} & Adv Loss Start Iter & 60000 \\
\multirow{15}{*}{} & Noise $\tau$ & 0.8 \\
\multirow{15}{*}{} & Pixel Count per Image & 37000$\sim$75500 \\
\multirow{15}{*}{} & Aspect Ratio & 0.7$\sim$1.8 \\
\multirow{15}{*}{} & Image per GPU & 6$\sim$18 \\
\multirow{15}{*}{} & Mask Probability & 0.1 \\
\multirow{15}{*}{} & Mask Ratio & 0.1$\sim$0.6 \\
\bottomrule
\end{tabular}}
\caption{\textbf{Details of 3DRAE architecture and hyperparameters.}}
\label{tab:supl:3drae_detail}
\vspace{-10pt}
\end{table}

\subsection{Frozen 2D Encoder}
For any given frozen 2D representation encoder, we use its pre-trained ViT-Base model and follow the protocols presented in RAE\cite{rae} if not otherwise specified.
In particular, we discard any [CLS] or [REG] token produced by 2D encoder, and keep all of the patch tokens. Notably, since our 3D latent representation is not directly derived from 2D encoder but processed and output by our Latent Fuse Neck, we do not have to ensure the tokens of 2D encoder to have zero mean and unit variance. Thus we just keep the original layer normalization in ViT architecture at the end of 2D encoder.

{\noindent}During training, we adopt dynamic image resolutions while ensuring that both height and width are divisible by 14. Since different 2D encoders employ varying patch sizes (e.g., 14 for DINOv2\cite{dinov2} and DA3\cite{da3}, or 16 for SigLIP2\cite{siglip2}), we resize all input images to a common resolution divisible by 14 before feeding them into the 2D encoder. This ensures that the resulting token counts are consistent with those produced under a patch size of 14, enabling uniform processing across different encoder backbones. For example, for an image at resolution 336$\times$296 that produces 336$\times$296/14/14=504 tokens under a patch size of 14, we resize it to (336/14*16)$\times$(296/14*16) resolution for SigLIP2 backbone, or keep the original 336$\times$296 resolution for DINOv2 and DA3 backbones.

\subsection{Latent Fuse Neck}
Our Latent Fuse Neck consists of 12 transformer layers that process the concatenation of multi-view patch tokens and 3D latent tokens. Through global self attention\cite{vggt}, it enables thorough information exchange between these tokens, effectively fusing multi-view information into the 3D latent tokens.
Following the practice in RAE\cite{rae}, we apply layer normalization with no learned affine parameter at the end of Latent Fuse Neck to ensure each 3D latent token has zero mean and unit variance. After the layer normalization, we apply batch normalization with mean and variance globally computed on ImageNet dataset\cite{imagenet} to normalize the distribution of our 3D latent representation.

\subsection{Latent Query Decoder}
Our Latent Query Decoder consists of 16 transformer layers that process the concatenation of ray map tokens and multi-view fused 3D latent tokens. Through global self attention\cite{vggt}, target-view information in the multi-view fused 3D latent tokens can be queried and retrieved by the ray map tokens to predict target-view images.
We extract ray map tokens for each target view from the output of the final transformer layer. These tokens, which encode necessary target view information, are then reshaped via unpatchification with a patch size of 14 to produce high-resolution images.
For optional point map output, we adopt a separate Latent Query Decoder with DPT head\cite{dpt} aggregating features from the 3rd, the 7th, the 11th and the 15th transformer layers for dense prediction.
Since our 3D latent representation is decoupled from the input multi-view images, we can query it with ray maps of arbitrary views to render the corresponding images and point maps.

To enhance the decoder's generalization to the deviated latent space of diffusion model, we follow~\cite{rae} to augment our 3D latent tokens with additive noise during training for robust decoding.

\begin{figure}[htbp]
\centering
\vspace{0pt}
\includegraphics[width=0.9\linewidth]{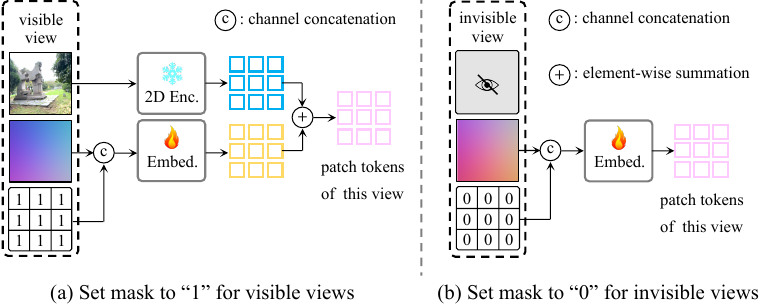}
\vspace{-5pt}
\caption{\textbf{Details of View Masking.}}
\label{fig:mask}
\vspace{-10pt}
\end{figure}
\subsection{Random Masking Strategy}
\label{randmask}
To improve the robustness of our 3DRAE to missing views and enable compatibility with diverse conditioning configurations (i.e., varying numbers of visible views), we introduce a random masking strategy during training of 3DRAE. Specifically, we randomly mask out a set of multi-view images with probability 0.1, and for each masked sample, we randomly set 60$\%$ to 90$\%$ of the views as invisible. As shown in Fig.\ref{fig:mask}, visibility is indicated by binary masks: all-ones for visible views, all-zeros for invisible views. For a visible view, we concatenate an all-ones mask with its Plücker ray map and feed them into the ray map embedding module. The resulting ray map tokens—now encoding visibility—are summed with patch tokens from the frozen 2D encoder to produce the final patch tokens. For an invisible view, we concatenate an all-zeros mask with its ray map; the resulting ray map tokens encode only pose information and directly serve as the final patch tokens. With a 3DRAE trained under this masking strategy, we can encode camera poses from a complete trajectory alongside partially observed images. The resulting conditioning signals thus contain full spatial extent information, requiring the 3DDiT to generate only the spatially-consistent appearance of invisible views.

\subsection{Impact of Adversarial Loss}
\begin{figure}[!ht]
\centering
\vspace{0pt}
\includegraphics[width=0.8\linewidth]{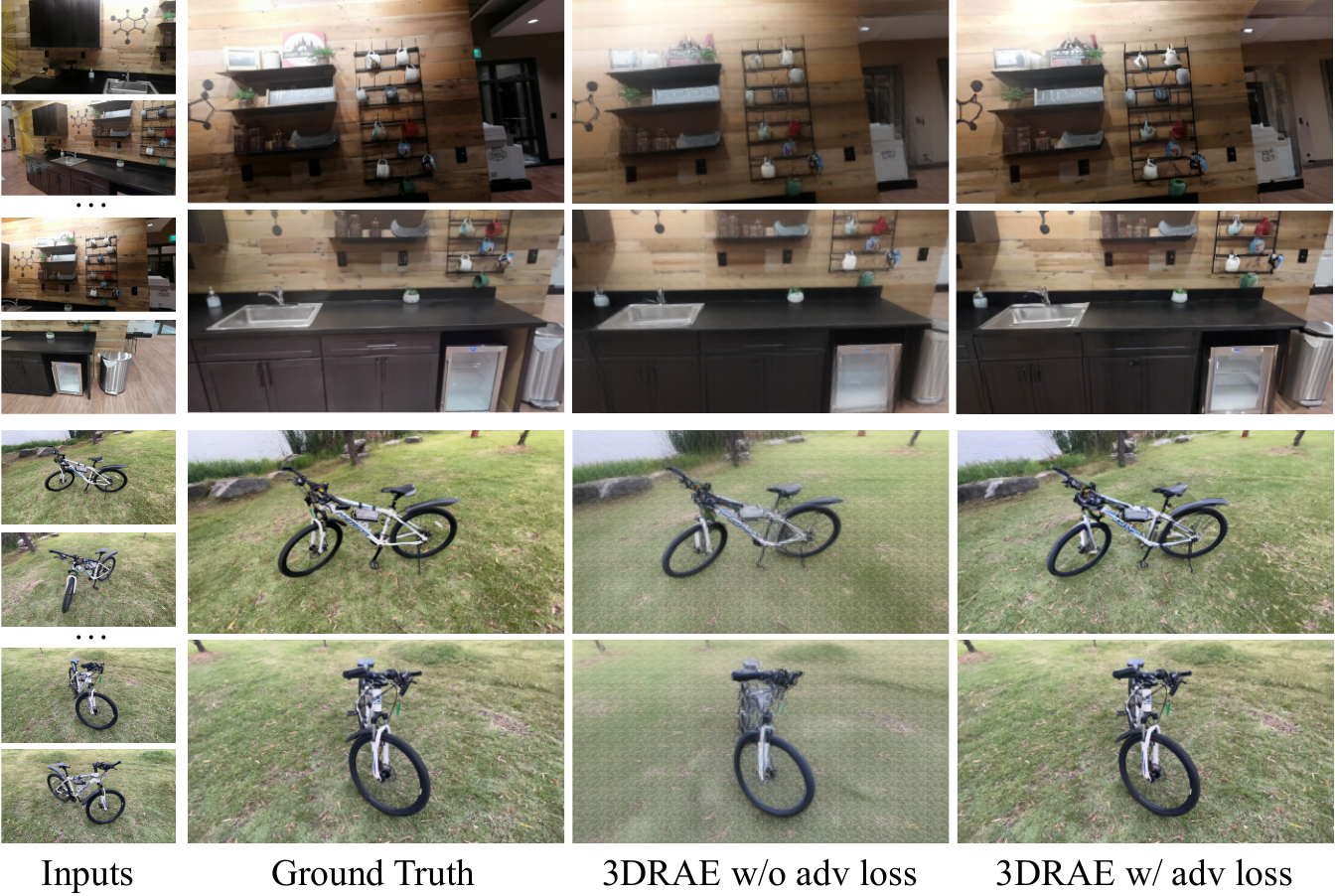}
\vspace{-5pt}
\caption{\textbf{Impact of Adversarial Loss.}}
\label{fig:adv}
\vspace{-10pt}
\end{figure}

We incorporate adversarial loss into our 3DRAE training. A discriminator, initialized from DINOv2-Small\cite{dinov2}, is tasked with distinguishing between images produced by our decoder and real images, enforcing alignment between the predicted and ground-truth image distributions.
Since such adversarial loss is commonly used in VAE training but has not been previously explored in novel view synthesis, we ablate its effect on rendering quality in Fig.\ref{fig:adv}. Interestingly, we observe that incorporating the adversarial loss leads to noticeable improvements in image quality, particularly when the target view differs substantially from the input views in terms of pose. We posit that the adversarial loss enhances the robustness of image rendering to pose variations.

\section{3DDiT Implementation}
\subsection{Diffusion Transformer Model}
We use LightningDiT\cite{lightningdit} as the backbone of our 3DDiT model by default.
Following \cite{rae}, we append a shallow but wide DDT head\cite{ddt} to the LightningDiT backbone for high-dimensional 3D latent denoising.
A continuous time schedule with time step restricted to real values in $[0,1]$ is employed for flow matching formulation, where the time step embedding is obtained from a Gaussian Fourier embedding layer.
We also adopt a dimension-dependent noise schedule shift\cite{sd3} during training.
For a schedule $t_n\in[0,1]$ and 3D latent token count $N$ with dimension $M$, the shifted timestep is calculated as $t_m=\frac{\alpha t_n}{1+(\alpha-1)t_n}$ where $\alpha=\sqrt{N{\times}M/D}$ is a dimention-dependent scaling factor. We follow \cite{sd3} to set $D=4096$ as the base dimension and $N{\times}M$ denotes the effective data dimension of our 3D latent space.

Detailed architecture and hyperparameter settings for 3DDiT at each model size are shown in Table \ref{tab:supl:3ddit_detail}.
In general, our 3DDiT is firstly trained for 80k steps on ImageNet\cite{imagenet}, and further trained for 80k steps on the same collection of multi-view datasets as the training of 3DRAE.
The difference is that, to enable our 3DDiT to support diverse multi-view conditioning configurations (i.e., unconditioning with number of views = 0, single-view conditioning with number of views = 1, sparse-view conditioning with number of views > 1), we set the masking probability to 1.0 to ensure we always use partial observed views as condition and set the masking ratio ranging from 0.6 to 0.9 to vary the number of observed views in condition.
To enable complete unconditioning with no observed view, we randomly drop all views and only keep the poses at a probability 0.1.
During the pre-training stage, such unconditioning is achieved by randomly drop class label conditions at a probability 0.1.
The class label embedding layer is dropped after the pre-training stage.

\begin{table}[ht!]
\vspace{-10pt}
\centering
\resizebox{0.8\linewidth}{!}{
\begin{tabular}{lll}
\toprule
\multicolumn{3}{c}{\textbf{(a) Network Architecture}} \\ \hline
\specialrule{0em}{1pt}{0pt}
\multirow{12}{*}{\textbf{LightningDiT Backbone\cite{lightningdit}}} & \multicolumn{2}{l}{\textbf{3DDiT-S}} \qquad\qquad\qquad\qquad\qquad\qquad\\
\multirow{12}{*}{} & Depth\qquad\qquad\qquad\qquad\qquad\qquad & 12 \\ 
\multirow{12}{*}{} & Hidden Dim & 384 \\ 
\multirow{12}{*}{} & Num Heads & 6 \\
\cmidrule(r){2-3}
\multirow{12}{*}{} & \multicolumn{2}{l}{\textbf{3DDiT-B}} \\
\multirow{12}{*}{} & Depth\qquad\qquad\qquad\qquad\qquad\qquad & 12 \\ 
\multirow{12}{*}{} & Hidden Dim & 768 \\ 
\multirow{12}{*}{} & Num Heads & 12 \\
\cmidrule(r){2-3}
\multirow{12}{*}{} & \multicolumn{2}{l}{\textbf{3DDiT-XL}} \\
\multirow{12}{*}{} & Depth\qquad\qquad\qquad\qquad\qquad\qquad & 28 \\ 
\multirow{12}{*}{} & Hidden Dim & 1152 \\ 
\multirow{12}{*}{} & Num Heads & 16 \\
\hline
\specialrule{0em}{1pt}{0pt}
\multirow{3}{*}{\textbf{DDT Head\cite{ddt}}} & Depth & 2\\
\multirow{3}{*}{} & Hidden Dim (Width) & 2048 \\
\multirow{3}{*}{} & Num Heads & 16 \\
\hline
\toprule
\specialrule{0em}{3pt}{0pt}
\multicolumn{3}{c}{\textbf{(b) Hyperparameters}} \\
\specialrule{0em}{1pt}{0pt} \hline
\specialrule{0em}{1pt}{1pt}
\multirow{14}{*}{\textbf{Training Details}} & lr schedule & Cosine Decay \\
\multirow{15}{*}{} & lr & 2e-4 \\
\multirow{14}{*}{} & Warmup Steps & 8000 \\
\multirow{14}{*}{} & Optimizer & AdamW \\
\multirow{14}{*}{} & $\beta_1$, $\beta_2$ & 0.9, 0.95 \\
\multirow{14}{*}{} & Ema Decay & 0.9995 \\
\multirow{14}{*}{} & Gradient Clip & 1.0 \\
\multirow{14}{*}{} & Iterations & 80000 \\
\multirow{14}{*}{} & Condition Drop Probability & 0.1 \\
\multirow{14}{*}{} & Pixel Count per Image & 37000$\sim$75500 \\
\multirow{14}{*}{} & Aspect Ratio & 0.7$\sim$1.8 \\
\multirow{14}{*}{} & Image per GPU & 6$\sim$18 \\
\multirow{14}{*}{} & Mask Probability & 1.0 \\
\multirow{14}{*}{} & Mask Ratio & 0.6$\sim$0.9 \\
\bottomrule
\end{tabular}}
\caption{\textbf{Details of 3DDiT architecture and hyperparameters.}}
\label{tab:supl:3ddit_detail}
\vspace{-10pt}
\end{table}

\subsection{Sampling Details}
During sampling of 3DDiT, we use standard ODE sampling with Euler sampler and 50 steps by default.
We also use classifier-free guidance (CFG) during sampling to balance diversity and fidelity.
Specifically, for unconditional prediction required by CFG, we directly drop the observed view images but keep the corresponding poses to ensure the conditioning latent tokens obtained from 3DRAE only contain the spatial extent information with no appearance constraint.
For simplicity, we always set CFG scale to 2.0 during sampling for all conditioning configurations in our experiments.

\begin{table}[!ht]
\vspace{-10pt}
    \centering
    \setlength{\tabcolsep}{5pt}
    \footnotesize
    \resizebox{0.75\linewidth}{!}{
    \begin{tabular}{l c c c c}
        \specialrule{0.12em}{0em}{0em}
        Name & Domain & \#Scenes & \#Frames & Weight \\
        \hline
        CO3D\cite{co3d} & Indoor/Object & 27K & 2.6M & 7.1\%\\
        WildRGBD\cite{wildrgbd} & Indoor/Object & 19k & 6.8M & 5.1\%\\
        ARKitScenes\cite{arkit} & Indoor/Scene & 3.4K & 0.9M & 13.2\%\\
        ScanNet\cite{scannet} & Indoor/Scene & 1.2K & 1.9M & 6.3\%\\
        ScanNet\textbf{++}\cite{scannetpp} & Indoor/Scene & 855 & 0.7M & 3.4\%\\
        Taskonomy\cite{taskonomy} & Indoor/Scene & 537 & 4.6M & 4.2\%\\
        HM3D\cite{hm3d} & Indoor/Scene & 899 & 9.5M & 4.7\%\\
        Hypersim\cite{hypersim} & Indoor/Scene & 767 & 74K & 5.0\%\\
        RE10K\cite{re10k} & Indoor/Scene & 66k & 8.9M & 17.3\%\\
        Waymo\cite{waymo} & Outdoor/Driving & 798 & 0.8M & 6.3\%\\
        Mapillary\cite{mapillary} & Outdoor/Driving & 74 & 27K & 0.6\%\\
        Virtual KITTI\cite{vkitti} & Outdoor/Driving & 50 & 21K & 0.7\%\\
        MVS-Synth\cite{mvssynth} & Outdoor/Driving & 120 & 12K & 0.6\%\\
        DL3DV\cite{dl3dv} & Outdoor/Scene & 6k & 2.2M & 8.4\%\\
        BlendedMVG\cite{blended} & In-the-wild/Scene & 501 & 115K & 5.3\%\\
        MegaDepth\cite{megadepth} & In-the-wild/Scene & 213 & 123K & 0.6\%\\
        TartanAir\cite{tartanair} & In-the-wild/Scene & 369 & 307K & 3.9\%\\
        ImageNet\cite{imagenet} & In-the-wild/Scene & 1.3M & 1.3M & 8.4\%\\
        \specialrule{0.12em}{0em}{0em}
    \end{tabular}
    }
    \caption{\textbf{Training Datasets.}}
    \label{tab:supl:datasets}
\vspace{-10pt}
\end{table}
\section{Training Datasets}
\label{datasets}
We preprocess the raw datasets listed in Table \ref{tab:supl:datasets} to train our 3DRAE and 3DDiT models. This collection covers a wide range of scenarios including indoor, outdoor, in-the-wild, object-centric and general scenes, enabling us to scale our method to large-scale data.
To balance domain variance across datasets, we assign each dataset a sampling weight that determines its probability of being sampled during training, independent of the number of scenes and frames it contains.
The preprocessing of these datasets follows the protocol presented in the official implementation of CUT3R\cite{cut3r} if not otherwise specified.
For RE10K\cite{re10k} without ground-truth depth, we directly disable the point map supervision during training.
For single-image data in ImageNet\cite{imagenet}, which has no pose annotation, we leverage MoGe2\cite{moge2} to estimate camera intrinsics for each image.
Given a set of multi-view images, we align all camera poses to the coordinate frame of the first view, which serves as the world frame. The distance of the farthest camera from the first view is then used as a scale factor to normalize all poses, ensuring consistent scale across different datasets.
For each data sample, all the images are cropped and resized to a random combination of resolution and aspect ratio with the pixel count constrained to a predefined range.
The number of multi-view images per data sample is randomly selected from a predefined range. Based on the number of views and pixel count, we dynamically adjust the batch size to maintain a stable total number of tokens to be processed.
Please see Table \ref{tab:supl:3drae_detail} and Table \ref{tab:supl:3ddit_detail} for specific hyperparameter settings.

\section{Benchmark Details}
To comprehensively evaluate the performance of our method, we construct a benchmark comprising 358 scenes from DL3DV\cite{dl3dv}, RE10K\cite{re10k}, CO3D\cite{co3d} and ScanNet\cite{scannet}, covering diverse scene types including indoor, outdoor, object-level, and general scenarios.
All the scenes are unseen during training, and are randomly selected subject to the constraint that the camera rotation and translation variation from the first to the last view exceeds an empirical threshold. This ensures substantial camera motion for thoroughly assessing the models' generative capabilities.
For each scene, we uniformly downsample the original video or image sequence to 16 views, which serves as the minimum evaluation unit.
To accommodate different conditioning configurations with varying numbers of observed views, we further sample from the 16 views accordingly. For single-view conditioning, we take only the first view as the observed view. For sparse-view conditioning, we uniformly sample observed views from the 16 views. All the 16 views excluding the observed ones are used for evaluation.
In our evaluation, we employ multiple metrics. Photometric metrics including PSNR and LPIPS are used to measure reconstruction quality. Image distribution metric FID assesses generated image quality. To evaluate spatial consistency, we adopt pose accuracy metrics following prior work\cite{viewcrafter}. Specifically, $\text{ATE}_\text{r}$ and $\text{ATE}_\text{t}$ compute rotation and translation errors between poses of generated multi-view images and ground-truth poses, based on SLAM evaluation protocols\cite{slambench,slambench_2}. The poses of generated images are estimated using VGGT\cite{vggt} with bundle adjustment\cite{ba}.

\section{Video Comparison on Long-Trajectory Generation}
Although our method primarily focuses on generating a queryable 3D latent representation of appearance and geometry across different poses, rather than producing videos composed of dense frame images, our 3DRAE decoder nevertheless support long-trajectory video rendering by querying our generated 3D latent representation at varying poses along a long trajectory.

{\noindent}In the supplementary video file \blue{\texttt{video\_compare.mp4}}, we present video results for long-trajectory (> 100 frames) generation alongside a comparison with baseline methods. As can be seen from the video, whereas 2D diffusion-based approaches struggle to render spatially consistent videos across different poses, our method accurately aligns with varying poses along a long trajectory.
Taking SEVA~\cite{seva} as an example, such approaches typically divide a long target trajectory into multiple chunks, generate each chunk separately, and then concatenate the results. This inevitably introduces spatial inconsistencies across chunks: the scene appearance generated in the current chunk changes in the next chunk, and the poses often exhibit abrupt jumps.
Moreover, compared to 2D diffusion-based methods that only generate frame images, our method simultaneously obtains a point map for each pose, enabling the construction of a scene point cloud and thus achieving truly 3D-level scene generation.
In comparison with regression-based methods, approaches such as LVSM~\cite{lvsm} and DepthSplat~\cite{depthsplat} are unable to complete unseen scene appearance without generative modeling. Consequently, when the pose undergoes large changes, the rendered images exhibit drift or collapse. In contrast, our method can generate plausible appearance for unseen regions.

{\noindent}We also note that, for long-trajectory video generation, 2D diffusion-based methods require per-frame denoising, which is highly time-consuming. In contrast, our method generates the 3D latent tokens of the scene in a single pass. When rendering video frames, no heavy denoising is needed, where we simply query our 3DRAE decoder to predict the image for each target pose, incurring a computational cost comparable to that of regression-based approaches. Therefore, compared to the task that only generate a small number of novel-view images, our approach offers a significant efficiency advantage in long-trajectory video generation.
Specifically, in Table~\ref{tab:video_time}, we report the average per-frame inference time when generating a 100-frame long-trajectory video, and find that our method is significantly faster than 2D diffusion-based methods (e.g., MotionCtrl~\cite{motionctrl}, ViewCrafter~\cite{viewcrafter}, SEVA~\cite{seva}) while being comparable to regression-based approaches (e.g., LVSM~\cite{lvsm}, DepthSplat~\cite{depthsplat}).

\begin{table*}[htbp]
    \centering
    \resizebox{0.99\linewidth}{!}{
    \begin{tabular}{c c c c c c c}
    \toprule
    & MotionCtrl~\cite{motionctrl} & ViewCrafter~\cite{viewcrafter} & SEVA~\cite{seva} & DepthSplat~\cite{depthsplat} & LVSM~\cite{lvsm} & Ours\\
    \midrule
    sec./frame & 0.738 & 12.638 & 5.325 & 0.010 & 0.013 & 0.032 \\
    \bottomrule
    \end{tabular}
    }
    \caption{\textbf{Per-frame inference time comparison for 100-frame long-trajectory video generation.}}
    \label{tab:video_time}
\end{table*}

\section{Visual Results}
In the supplementary video file \blue{\texttt{3D\_gen.mp4}}, we include additional 3D scene generation results under diverse conditioning configurations, as well as zero-shot results on unseen domains. The details are as follows.
\subsection{Additional 3D Scene Generation Results}
\begin{figure}[!ht]
\centering
\vspace{0pt}
\includegraphics[width=0.95\linewidth]{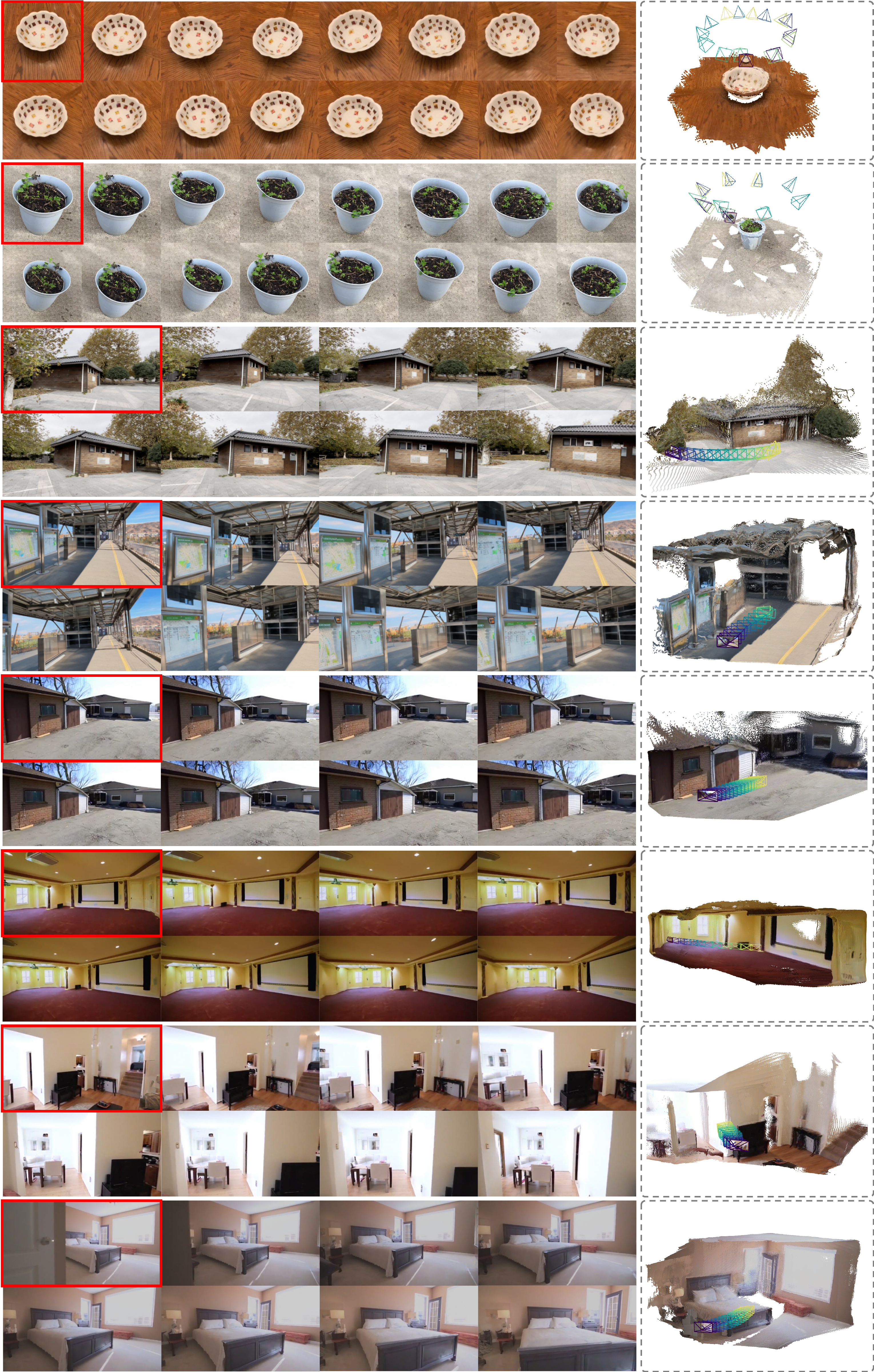}
\vspace{-0pt}
\caption{\textbf{Generated 3D Scenes under Single-View Conditioning.} The red bounding boxes denote the conditional views. Generated point clouds are shown on the right.}
\label{fig:p1_vis}
\vspace{-0pt}
\end{figure}

\noindent\textbf{Single-View Conditioning.}
\addcontentsline{toc}{subsubsection}{Single-View Conditioning}
In Fig.\ref{fig:p1_vis}, we present extensive results of single-view conditioned 3D scene generation. For each sample, we provide 3DDiT with a single image as the conditional observation, along with a variable-length pose sequence specifying the spatial extent. As demonstrated, our method performs remarkably well on both object-level (rows 1$\sim$2), outdoor scene-level (rows 3$\sim$5) and indoor scene-level (rows 6$\sim$8) 3D generation. It supports pose sequences of varying lengths and decodes images at different resolutions and aspect ratios, highlighting the advantage of our 3DRAE: the decoupling of 3D latent representation from multi-view observations.

\begin{figure}[!ht]
\centering
\vspace{0pt}
\includegraphics[width=0.77\linewidth]{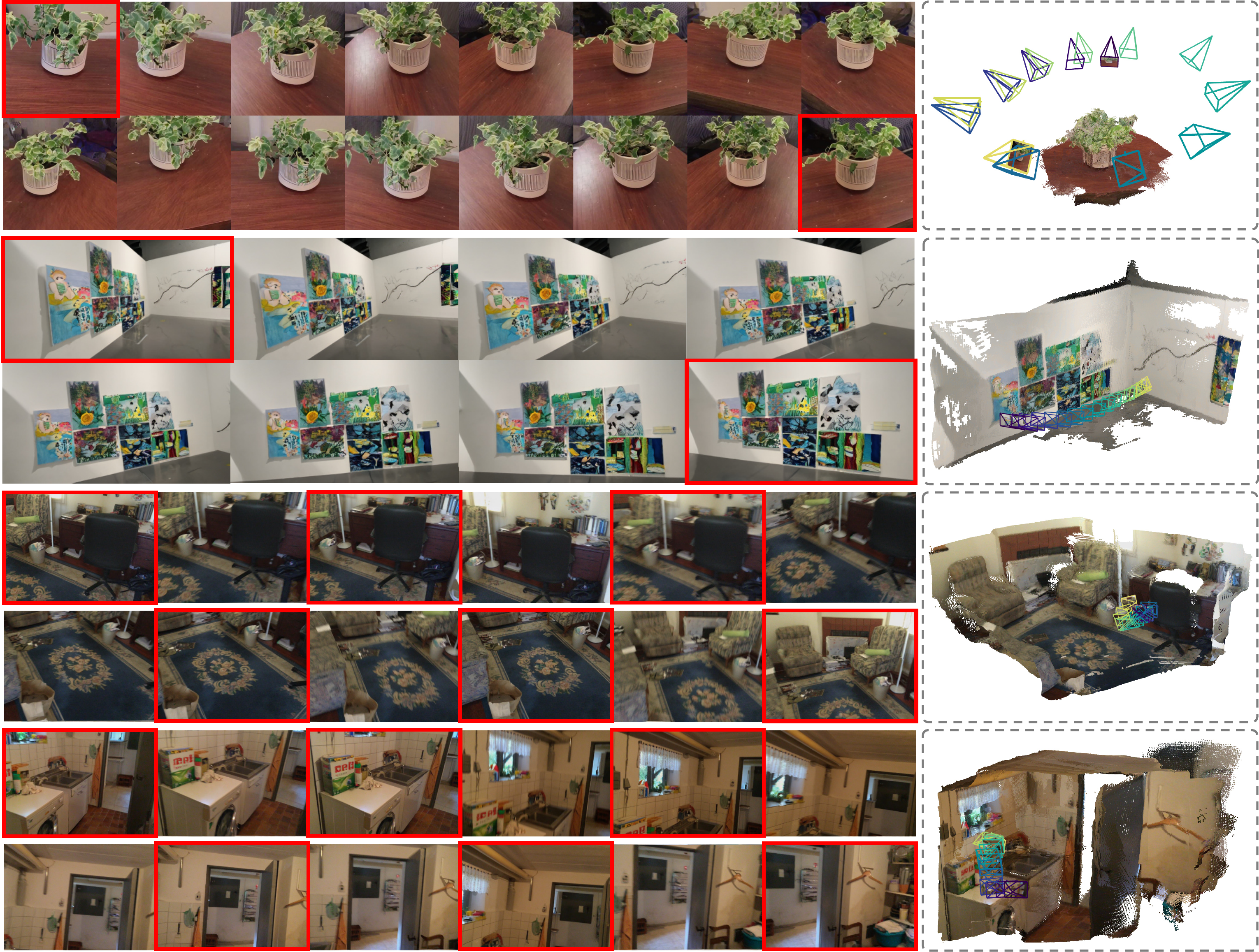}
\vspace{-5pt}
\caption{\textbf{Generated 3D Scenes under Sparse-View Conditioning.} The red bounding boxes denote the conditional views. Generated point clouds are shown on the right.}
\label{fig:pn_vis}
\vspace{-5pt}
\end{figure}

\noindent\textbf{Sparse-View Conditioning.}
\addcontentsline{toc}{subsubsection}{Sparse-View Conditioning}
In Fig.\ref{fig:pn_vis}, we present extensive results of sparse-view conditioned (i.e., number of views > 1) 3D scene generation.
As shown in the figure, our method consistently generates plausible 3D scenes under varying numbers of conditional views (rows 1$\sim$2 with 2 views and rows 3$\sim$4 with 6 views). This demonstrates its highly competitive performance even in sparse-view settings where reconstruction fidelity is prioritized.

\begin{figure}[!h]
\centering
\vspace{0pt}
\includegraphics[width=0.77\linewidth]{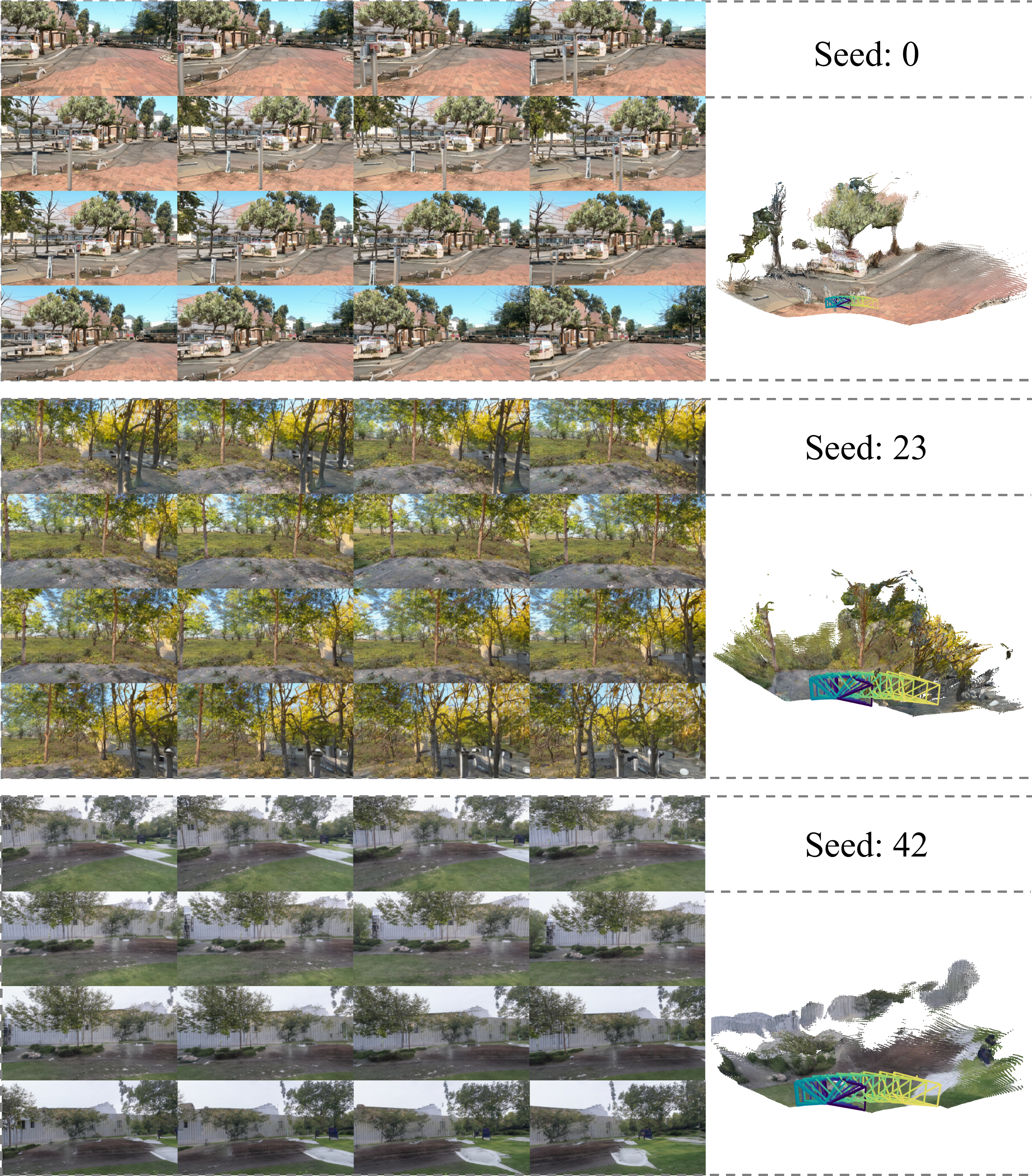}
\vspace{-5pt}
\caption{\textbf{Generated 3D Scenes without Appearance Conditioning.} Using the same camera trajectory condition with different random seeds, we generate 3D scenes with varied appearances.}
\label{fig:unc_vis}
\vspace{-10pt}
\end{figure}

\noindent\textbf{Unconditioning.}
\addcontentsline{toc}{subsubsection}{Unconditioning}
In Fig.\ref{fig:unc_vis}, we further show the results of unconditional 3D scene generation where we only provide 3DDiT with only camera trajectory as input condition but no image observation.
By varying the random seed, we can generate scenes with diverse appearances within the same spatial scope defined by a fixed camera trajectory. This demonstrates that our trained 3DDiT possesses the capability to synthesize general scene appearances even in the absence of image observations.

\subsection{Zero-Shot Results}
\begin{figure}[!ht]
\centering
\vspace{0pt}
\includegraphics[width=0.95\linewidth]{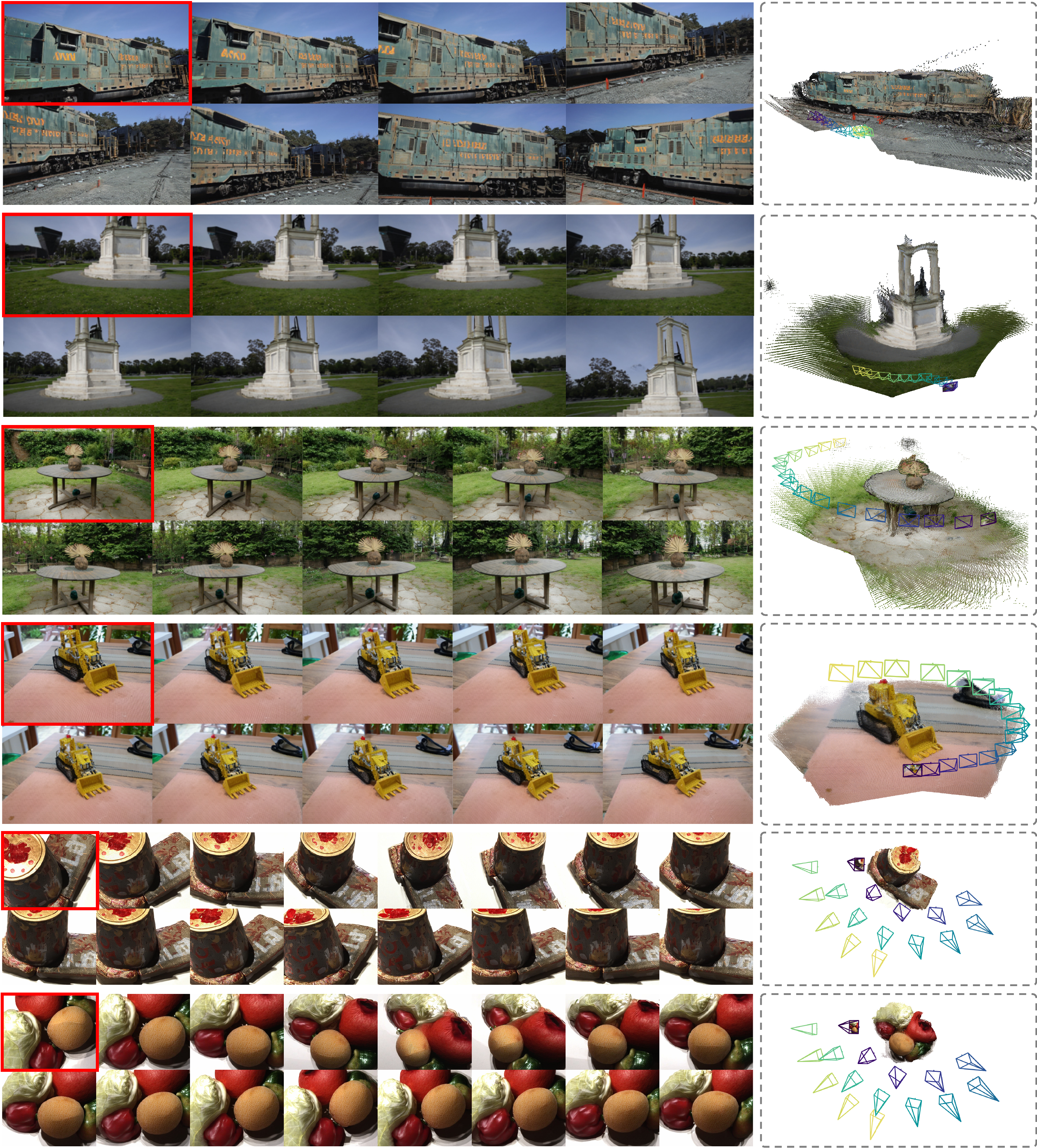}
\vspace{-5pt}
\caption{\textbf{Zero-Shot Generated 3D Scenes under Single-View Conditioning.} \textbf{Rows 1$\sim$2:} results on Tanks and Temples dataset\cite{tnt}. \textbf{Rows 3$\sim$4:} results on Mip-NeRF 360 dataset\cite{mip360}. \textbf{Rows 5$\sim$6}: results on DTU dataset\cite{dtu}. The red bounding boxes denote the conditional views. Generated point clouds are shown on the right.}
\label{fig:zeroshot_vis}
\vspace{-10pt}
\end{figure}

As shown in Fig.\ref{fig:zeroshot_vis}, we further evaluate the zero-shot generalization capability of our approach on three datasets unseen during training: Tanks and Temples\cite{tnt}, Mip-NeRF 360\cite{mip360}, and DTU\cite{dtu}. These span indoor, outdoor, and object-centric scenes, presenting diverse out-of-domain challenges. Despite not being trained on web-scale data like multi-view or video diffusion models, our method exhibits strong generalization performance across diverse domains, demonstrating its potential for real-world applications.

\section{3DRAE under Extremely Dense Multi-View Observation}
\begin{figure}[!ht]
\centering
\vspace{0pt}
\includegraphics[width=0.9\linewidth]{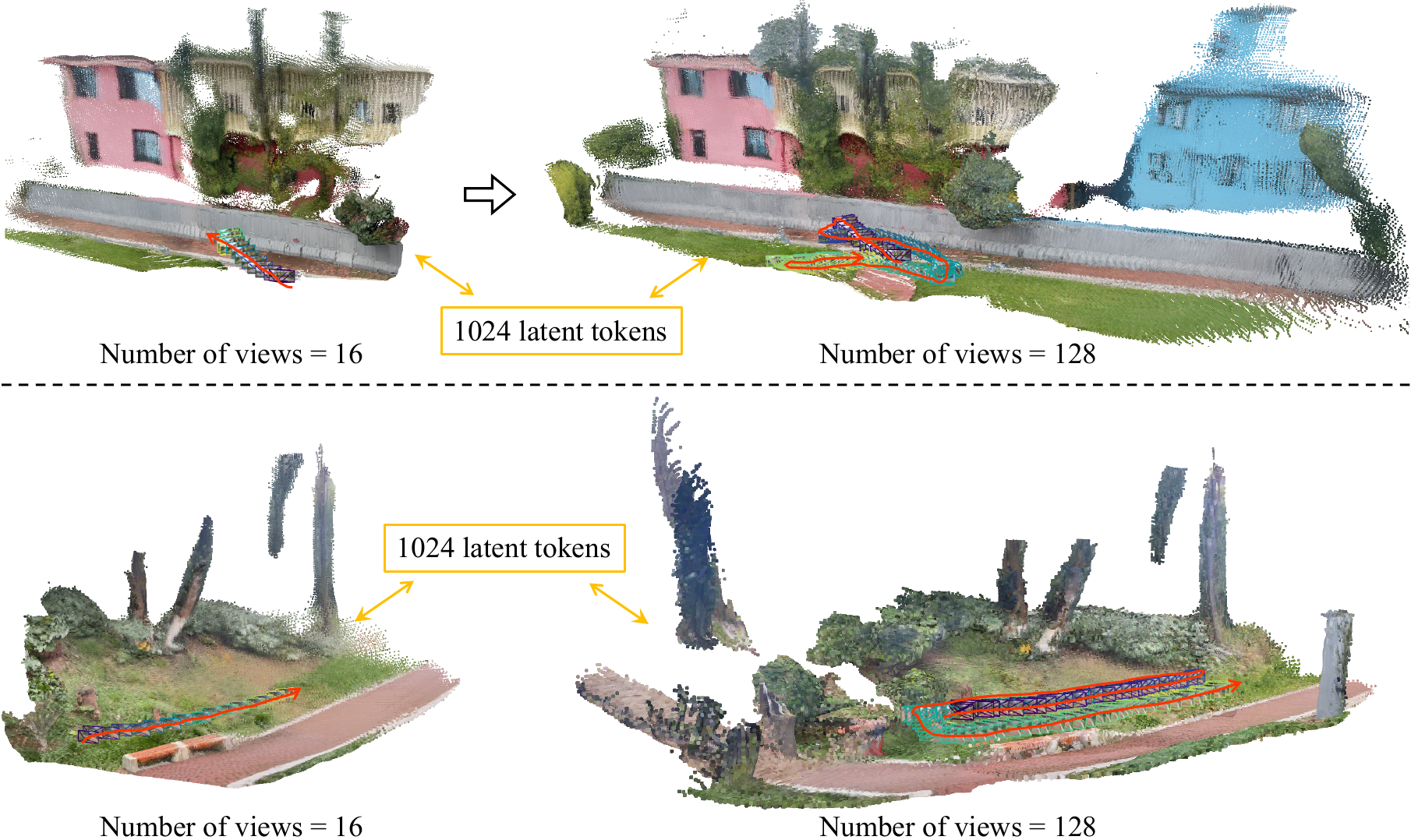}
\vspace{-5pt}
\caption{\textbf{Representing Densely Observed 3D Scenes using 3DRAE.} Our 3DRAE can represent the same 3D scenes but observed by different numbers of views (16 vs. 128 views) using the same set of 1K tokens. The red arrows highlight conditioning camera trajectories that specify the spatial extents of the generated scenes.}
\label{fig:dense_vis}
\vspace{-10pt}
\end{figure}

Although our 3DRAE was trained with at most 18 views per scene to learn compact 3D latent representations, we investigate its behavior under significantly denser observational views in Fig.\ref{fig:dense_vis}.
To our surprise, our 3DRAE can represent extremely densely observed scenes with up to 128 views using only 1024 3D latent tokens, without degradation in reconstruction quality compared to the 16-view setting. This demonstrates the strong capability of our 3DRAE to generalize to large-scale scenes with many more observational views while maintaining fixed and compact representation complexity. Please refer to the supplementary video file \blue{\texttt{dense\_recon.mp4}} for better visualization.

\section{Broader Impacts}
\label{impact}
Our method introduces a novel paradigm: repurposing 2D representation models to derive 3D-grounded latent representation for 3D scenes. Such 3D latent representation preserves the rich semantics inherent in 2D models while being grounded with 3D awareness through supervision on novel view synthesis and point map reconstruction. Crucially, such compact 3D latent representation is substantially more efficient than representing 3D scenes as multi-view images or videos. By performing diffusion modeling directly within this 3D latent space at scale, we demonstrate efficient 3D scene generation with strong spatial consistency, validating the potential of this 3D latent representation as a powerful alternative to multi-view 2D latent representation commonly adopted in prior works\cite{seva,viewcrafter,motionctrl,prometheus,flexworld,vmem}.

More broadly, this 3D latent space, particularly when derived from SigLIP2, is naturally compatible with mainstream large multimodal models\cite{qwen25vl,qwen3vl}. Potentially, it enables the conversion of multi-view observed 3D scenes into a fixed length set of 3D-aware and language-aligned tokens, facilitating 3D scene understanding without relying on image- or video-based large multimodal models as in many existing approaches\cite{vlm3r,spatialmllm,spar7m,vgllm}.
This opens new possibilities toward building embodied agents that unify 3D worlds and language. Furthermore, our 3D latent space holds promise for unifying 3D scene understanding and generation within shared representation, contributing to the development of physical world models. These promising directions are worth pursuing in future studies.


\newpage
\clearpage
\section*{NeurIPS Paper Checklist}

The checklist is designed to encourage best practices for responsible machine learning research, addressing issues of reproducibility, transparency, research ethics, and societal impact. Do not remove the checklist: {\bf The papers not including the checklist will be desk rejected.} The checklist should follow the references and follow the (optional) supplemental material.  The checklist does NOT count towards the page limit. 

Please read the checklist guidelines carefully for information on how to answer these questions. For each question in the checklist:
\begin{itemize}
    \item You should answer \answerYes{}, \answerNo{}, or \answerNA{}.
    \item \answerNA{} means either that the question is Not Applicable for that particular paper or the relevant information is Not Available.
    \item Please provide a short (1--2 sentence) justification right after your answer (even for \answerNA). 
\end{itemize}

{\bf The checklist answers are an integral part of your paper submission.} They are visible to the reviewers, area chairs, senior area chairs, and ethics reviewers. You will also be asked to include it (after eventual revisions) with the final version of your paper, and its final version will be published with the paper.

The reviewers of your paper will be asked to use the checklist as one of the factors in their evaluation. While \answerYes{} is generally preferable to \answerNo{}, it is perfectly acceptable to answer \answerNo{} provided a proper justification is given (e.g., error bars are not reported because it would be too computationally expensive'' or ``we were unable to find the license for the dataset we used''). In general, answering \answerNo{} or \answerNA{} is not grounds for rejection. While the questions are phrased in a binary way, we acknowledge that the true answer is often more nuanced, so please just use your best judgment and write a justification to elaborate. All supporting evidence can appear either in the main paper or the supplemental material, provided in appendix. If you answer \answerYes{} to a question, in the justification please point to the section(s) where related material for the question can be found.

IMPORTANT, please:
\begin{itemize}
    \item {\bf Delete this instruction block, but keep the section heading ``NeurIPS Paper Checklist"},
    \item  {\bf Keep the checklist subsection headings, questions/answers and guidelines below.}
    \item {\bf Do not modify the questions and only use the provided macros for your answers}.
\end{itemize}


\begin{enumerate}

\item {\bf Claims}
    \item[] Question: Do the main claims made in the abstract and introduction accurately reflect the paper's contributions and scope?
    \item[] Answer: \answerYes{} 
    \item[] Justification: We have clearly stated the contributions and scope in the Abstract and Introuction.

\item {\bf Limitations}
    \item[] Question: Does the paper discuss the limitations of the work performed by the authors?
    \item[] Answer: \answerYes{} 
    \item[] Justification: We have discussed the limitations in Sec. \ref{limitation} of our Appendix.

\item {\bf Theory assumptions and proofs}
    \item[] Question: For each theoretical result, does the paper provide the full set of assumptions and a complete (and correct) proof?
    \item[] Answer: \answerYes{} 
    \item[] Justification: We established our assumptions and proofs upon papers cited.

    \item {\bf Experimental result reproducibility}
    \item[] Question: Does the paper fully disclose all the information needed to reproduce the main experimental results of the paper to the extent that it affects the main claims and/or conclusions of the paper (regardless of whether the code and data are provided or not)?
    \item[] Answer: \answerYes{} 
    \item[] Justification: All relevant details are provided within the text of our main paper and appendix, or through the references we have cited.

\item {\bf Open access to data and code}
    \item[] Question: Does the paper provide open access to the data and code, with sufficient instructions to faithfully reproduce the main experimental results, as described in supplemental material?
    \item[] Answer: \answerNo{} 
    \item[] Justification: We promise we will release all the data and code upon the acceptance of our paper.

\item {\bf Experimental setting/details}
    \item[] Question: Does the paper specify all the training and test details (e.g., data splits, hyperparameters, how they were chosen, type of optimizer) necessary to understand the results?
    \item[] Answer: \answerYes{} 
    \item[] Justification: All relevant details are provided within the text of our main paper and appendix, or through the references we have cited.

\item {\bf Experiment statistical significance}
    \item[] Question: Does the paper report error bars suitably and correctly defined or other appropriate information about the statistical significance of the experiments?
    \item[] Answer: \answerNo{} 
    \item[] Justification: No, due to the high computational cost of training large diffusion models on massive multi-view datasets, we only performed one final training run and evaluated on fixed test sets. However, all the quantitative results we reported in our experiments were averaged over three runs with different random seeds for the test-time sampling process.

\item {\bf Experiments compute resources}
    \item[] Question: For each experiment, does the paper provide sufficient information on the computer resources (type of compute workers, memory, time of execution) needed to reproduce the experiments?
    \item[] Answer: \answerYes{} 
    \item[] Justification: We have reported the computer resources we used in the experiments section.
    
\item {\bf Code of ethics}
    \item[] Question: Does the research conducted in the paper conform, in every respect, with the NeurIPS Code of Ethics \url{https://neurips.cc/public/EthicsGuidelines}?
    \item[] Answer: \answerYes{} 
    \item[] Justification: The research conducted in the paper conform the NeurIPS Code of Ethics.

\item {\bf Broader impacts}
    \item[] Question: Does the paper discuss both potential positive societal impacts and negative societal impacts of the work performed?
    \item[] Answer: \answerYes{} 
    \item[] Justification: We have discussed the potential impacts of our paper in Sec. \ref{impact} of our Appendix.
    
\item {\bf Safeguards}
    \item[] Question: Does the paper describe safeguards that have been put in place for responsible release of data or models that have a high risk for misuse (e.g., pre-trained language models, image generators, or scraped datasets)?
    \item[] Answer: \answerNA{} 
    \item[] Justification: The released model does not have a high risk for misuse.

\item {\bf Licenses for existing assets}
    \item[] Question: Are the creators or original owners of assets (e.g., code, data, models), used in the paper, properly credited and are the license and terms of use explicitly mentioned and properly respected?
    \item[] Answer: \answerYes{} 
    \item[] Justification: We have cited the open-source code and datasets mentioned in the paper and adhered to the terms of use.

\item {\bf New assets}
    \item[] Question: Are new assets introduced in the paper well documented and is the documentation provided alongside the assets?
    \item[] Answer: \answerYes{} 
    \item[] Justification: We are committed to releasing our dataset and code as soon as possible.

\item {\bf Crowdsourcing and research with human subjects}
    \item[] Question: For crowdsourcing experiments and research with human subjects, does the paper include the full text of instructions given to participants and screenshots, if applicable, as well as details about compensation (if any)? 
    \item[] Answer: \answerNA{} 
    \item[] Justification: This work does not involve crowdsourcing nor research with human subjects.

\item {\bf Institutional review board (IRB) approvals or equivalent for research with human subjects}
    \item[] Question: Does the paper describe potential risks incurred by study participants, whether such risks were disclosed to the subjects, and whether Institutional Review Board (IRB) approvals (or an equivalent approval/review based on the requirements of your country or institution) were obtained?
    \item[] Answer: \answerNA{} 
    \item[] Justification: This work does not involve crowdsourcing nor research with human subjects.

\item {\bf Declaration of LLM usage}
    \item[] Question: Does the paper describe the usage of LLMs if it is an important, original, or non-standard component of the core methods in this research? Note that if the LLM is used only for writing, editing, or formatting purposes and does \emph{not} impact the core methodology, scientific rigor, or originality of the research, declaration is not required.
    \item[] Answer: \answerNA{} 
    \item[] Justification: Method development in this research does not involve LLMs as any important, original, or non-standard components.

\end{enumerate}

\end{document}